\documentclass{article} % For LaTeX2e
\usepackage[final]{colm2026_conference}

\usepackage{microtype}
\usepackage{hyperref}
\usepackage{url}
\usepackage{booktabs}

\usepackage{graphicx}
\usepackage{subcaption}
\usepackage{float}
\usepackage{tcolorbox}
\tcbuselibrary{skins, breakable}
\usepackage{booktabs}
\usepackage{tabularx}
\usepackage{xcolor}
\usepackage{array}
\usepackage{multirow}
\usepackage{placeins}

\usepackage{lineno}

\definecolor{darkblue}{rgb}{0, 0, 0.5}
\hypersetup{colorlinks=true, citecolor=darkblue, linkcolor=darkblue, urlcolor=darkblue}

\usepackage{tikz}
\def\checkmark{\tikz\fill[scale=0.4](0,.35) -- (.25,0) -- (1,.7) -- (.25,.15) -- cycle;}

\title{Semantic Bandits: In-Context Exploration-Exploitation is Biased by Semantic Priors}

\author{
  David Eric Austin$^{1,2}$ \quad
  Kaheer Suleman$^{3}$ \quad
  Jackie Chi Kit Cheung$^{1,2,4}$ \quad \\[4pt]
  \small
  $^{1}$McGill University \\
  \small
  $^{2}$Mila -- Quebec AI Institute \\
  \small
  $^{3}$Skyfall AI \\
  \small
  $^{4}$Canada CIFAR AI Chair, Mila
}

\begin{document}

\ifcolmsubmission
\linenumbers
\fi

\maketitle

% LLM priors about actions and reward can cause significant, unexpected impacts on exploration behaviour in in-context multi-armed bandit tasks

\begin{abstract}
Large language models (LLMs) are increasingly deployed as decision-making agents
in settings that require sophisticated environmental exploration. However, existing work has raised questions about how LLMs actually balance exploration and exploitation. Unlike classical agents, LLM agents engage with tasks through natural language, exposing them to semantic information with no formal counterpart in the task structure. We introduce the \textit{semantic bandit}, an extension of the multi-armed bandit setting that explicitly considers the textual labels assigned to actions, and use it to study how \textit{semantic priors} --- inductive biases arising from associations between language and expected reward learned during pre-training, shape LLM exploration behaviour. We find that semantically informative action labels reduce exploration in favour of exploitation, improving performance when aligned with the reward structure and severely degrading it when misaligned. We further find that negative rewards trigger substantially more exploration than equivalent positive rewards, consistent with an expected-scale bias induced by reward conventions common in pre-training data. Overall, we argue that the use of language to define the environment and rewards introduces unavoidable biases derived from the fact that the model is trained on word co-occurence, with implications for the reliability and robustness of LLM agents in real-world decision-making settings.
\end{abstract}

\section{Introduction}
Large language models (LLMs) exhibit sophisticated capabilities on a wide variety of tasks \citep{bubeck2023sparks, srivastava2023beyond}. This has led to widespread deployment of LLM-based decision-making agents in in-context reinforcement learning (ICRL) tasks, which require acting on incomplete information and adapting across sequential turns based on feedback \citep{reflexion, zhouWebArenaRealisticWeb2023, drouinWorkArenaHowCapable2024}. Any such decision-making agent must balance between exploring new actions to gather information and exploiting past actions known to yield high reward \citep{suttonReinforcementLearningIntroduction2020}.  

While prior work \citep{krishna_icl, monea_bandit} evaluates ICRL using frameworks from classical reinforcement learning (RL), we argue that these frameworks are incomplete. Classical decision-making agents operate on symbolic problem representations. These representations are deliberately minimal: they preserve the formal structure of the task (states, actions, rewards) and discard everything else. In contrast, LLMs capture additional semantic information beyond the formal task structure. The name assigned to an action \citep{monea_bandit, min2022rethinkingroledemonstrationsmakes}, the magnitude and sign of numeric values \citep{mirzadeh2025gsmsymbolic}, and the domain in which a problem is situated \citep{kambhampatiPositionLLMsCant2024} are all semantically meaningful to a language model in ways they are not to a classical agent. This additional information may reflect genuine regularities in the world or it may consist of inscrutable associations that are not human-interpretable, learned as word co-occurrence statistics during pre-training. We refer to the inductive biases that arise from these learned associations as \textit{semantic priors}. These biases have no formal counterpart in classical RL.

\begin{figure*}[ht]
  \begin{center}
    \centerline{\includegraphics[width=\linewidth]{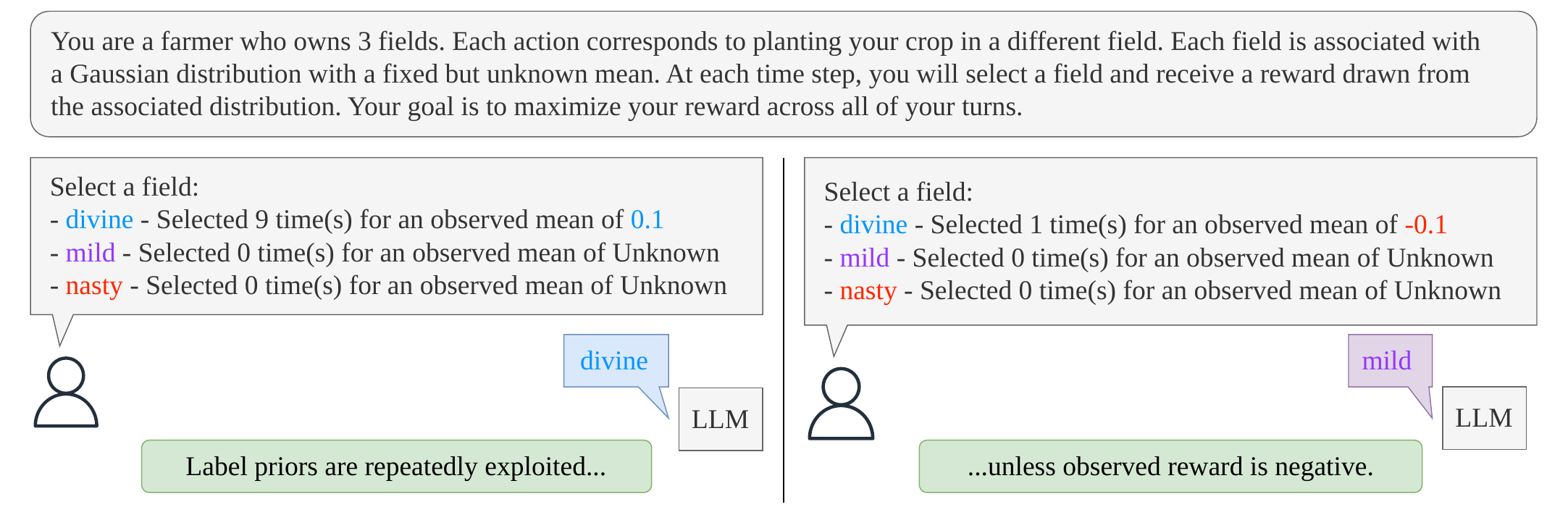}}
  \end{center}
  \caption{
      \textbf{LLM exploration in multi-armed bandit problems is systematically biased by associations between language and expected reward learned in pre-training.} In this example, the model is biased towards selecting actions with positive-sentiment labels ("divine"). It repeatedly exploits this action when the observed reward is positive (left), missing potentially higher-reward actions with neutral or negative sentiment labels. This label bias is nullified when the observed reward is negative, leading to significantly more exploration.
      }
    \label{fig:main_fig}
\end{figure*}

Due to this fundamental difference, evaluating ICRL using frameworks from classical RL leaves out crucial explanatory variables. To address this shortcoming, we introduce the semantic multi-armed bandit (Figure \ref{fig:main_fig}), a novel extension of the canonical bandit problem that explicitly parametrizes key aspects of the textual representation alongside the standard reward distributions, and use it to demonstrate how ICRL exploration is biased by semantic priors.

%%%%%%%%%% Main Figure
%%%%%%%%%% END

% To enhance LLMs for in-context exploration, we leverage known bandit algorithms such as Upper- Confidence Bound (UCB) algorithm, which have been proven "optimal" under mild conditions. - EVOLVE - Drop TS

We report three primary findings:
\begin{itemize}
    \item Action nomenclatures --- the textual labels assigned to bandit arms --- significantly influence exploration behaviour, even when they are not relevant to the task space. When the nomenclature is aligned with the underlying reward structure (e.g., positive-sentiment labels assigned to high-reward arms), models exploit semantic priors to achieve low regret with little exploration. When nomenclature is misaligned, the same priors misdirect the model toward low-reward actions, dramatically increasing cumulative regret.
    \item Reward polarity acts as a strong contextual cue: negative reward values trigger substantially more exploration than positive values, suggesting that LLMs interpret reward not as an abstract numerical signal but relative to an expected scale.
    \item These biases interact --- when the reward is negative, models are much less likely to exploit biases from action nomenclature (Figure \ref{fig:main_fig}).
\end{itemize}  

While the impact of semantic priors on classification and question answering tasks has been established \citep{jiang-etal-2024-peek, mirzadeh2025gsmsymbolic, cheng-etal-2025-stochastic, shojaee2025the}, their impact on decision-making tasks remains underexplored. \citet{monea_bandit} identified that semantic information plays a role in LLM exploration, but to our knowledge we are the first to characterize the potentially misleading semantic biases involved. We argue that the use of language as a representation space in ICRL tasks introduces semantic priors that are not derived from formal problem structure, leading to helpful or harmful effects depending on the alignment between the prior and the true reward distribution. 

Our most surprising finding is that the influence of these semantic priors is strong enough to nullify principled exploration behaviour in one of the simplest decision-making problems that can be constructed. As LLMs are increasingly deployed in real-world decision settings like recommendation systems and autonomous agents, understanding when semantic context aids adaptation and when it induces systematic errors is critical for safe and reliable deployment.

\section{Related work}
\paragraph{In-context reinforcement learning (ICRL).}
In-context learning is a paradigm in which pre-trained LLMs solve novel problems given a task description and a few examples in the prompt \citep{brown2020language, dong-etal-2024-survey}. \citet{min2022rethinkingroledemonstrationsmakes} argue that LLMs do not learn from in-context examples. Rather, the reasoning logic is learned implicitly in pre-training and adapted in-context to match the demonstration format.

We focus on in-context reinforcement learning (ICRL), a special class of in-context learning, in which LLMs act as decision-making agents --- adapting their performance entirely from in-context reward signal across sequential turns \citep{krishna_icl, monea_bandit, nie2025evolveevaluatingoptimizingllms}. While there is a significant body of work investigating LLM decision-making in application contexts \citep{reflexion, zhouWebArenaRealisticWeb2023, wangVoyagerOpenEndedEmbodied2023a}, fewer works have attempted to make claims about generalized decision-making concepts like exploration, planning, and generalization. We focus on exploration, which is traditionally assessed through the multi-armed bandit (MAB) problem \citep{suttonReinforcementLearningIntroduction2020}. In contrast to \citet{pan2025large}, we operationalize exploration through arm coverage, switching, and cumulative regret rather model-based analyses that separate directed exploration, random exploration, perseveration, and semantic-prior-driven choice. While some prior work has explored fine-tuning approaches \citep{nie2025evolveevaluatingoptimizingllms, icrl_algodist, iclr_supervised} for improving exploration behaviour, we follow \citet{krishna_icl}, who focus on identifying failure modes without fine-tuning. We also extend work by \citet{monea_bandit}, who identify that LLMs are slower to learn in ICRL tasks when semantic labels are removed. We extend this body of work by exploring the relationship between semantic information and ICRL behaviour. 

\paragraph{Reasoning in LLMs.}
LLMs have demonstrated impressive performance on reasoning benchmarks, prompting claims of generalized abstract reasoning capabilities \citep{bubeck2023sparks, weiemergent}. However, performance on logical and mathematical reasoning tasks remains highly sensitive to structure-preserving linguistic perturbations \citep{Shi2023LargeLM, jiang-etal-2024-peek, mirzadeh2025gsmsymbolic, tang2023large, cheng2025can}, potentially indicating probabilistic pattern-matching from immense, closed-source training corpora, rather than principled reasoning \citep{kambhampatiPositionLLMsCant2024, jiang-etal-2024-peek, mirzadeh2025gsmsymbolic, zhang2023paradox} . \citet{tang2023large} argue that LLM reasoning is fundamentally semantic, unlike in human reasoning, which is related to but distinct from language \citep{mahowaldDissociatingLanguageThought2024a}. Most prior work has focused on classification and question answering task. We extend this by exploring how these effects manifest in ICRL settings.

\section{Problem definition}
\paragraph{Classical reinforcement learning} 
We define an environment as the tuple $\{S, A, R, T\}$, where $S$ is the state space, $A$ is the action set, $R : A \rightarrow \Delta(\mathbb{R})$ is the reward function and $T$ is the episode length.\footnote{As the MAB is single-state, we simplify the Markov decision process formulation by removing the state transition function and discount factor. Also, $\Delta$ denotes the probability simplex.} A classical agent operates on a symbolic rendering generated by $\rho_{\Sigma} : S \times A \times T \rightarrow \Sigma$, where $\Sigma$ is an abstract representation space designed to preserve only formal structure. Its policy $\pi_{\Sigma} : \Sigma \rightarrow A$ is invariant to everything outside this structure.

\paragraph{Classical multi-armed bandit (MAB)}
The canonical problem for measuring exploration is the multi-armed bandit (MAB). A MAB instance has a single state and is fully defined by a set of $k$ arms $A = [a_1, \ldots, a_k]$, each associated with an unknown reward distribution $r_i$. Over $T$ turns, the agent selects an arm and observes the resulting reward, aiming to maximize cumulative reward. Since the distributions are unknown, the agent must balance exploration of the distributions with exploitation of promising arms.

\paragraph{LLMs as decision-making agents}
Unlike the symbolic agent, an LLM-based agent $M$ operates on a \textit{textual} rendering \citep{tang2023large} generated by $\rho_{V^*} : S \times A \times T \times C \rightarrow V^*$, where $V^*$ is the space of all token sequences and $C$ is the set of scenarios. This rendering includes aspects of how the environment is rendered in language as well as additional instructions and prompting details. Its policy $\pi_M : V^* \rightarrow V^*$ maps a linguistic problem description to a linguistic action. This output is passed to a parsing function $\phi : V^* \rightarrow A$ which converts the linguistic output to a concrete action. Crucially, $\rho_{V^*}$ and $\pi_M$ encode information that is not derived from $\{S, A, R, T\}$, such as the semantic associations about the scenario or between action labels and real-world outcomes. This information is legible to $M$ as semantic priors. Symbolic agents are blind to it by construction; LLM agents are not. This offers a potential advantage for LLM agents when the semantic prior aligns with the true task structure.

\paragraph{The semantic bandit}
\label{sec:sem_bandit_def}

While a classical agent engages only with the formal MAB instance, an LLM agent engages with a natural language description of it. This means the problem is no longer fully specified the reward distributions $[r_1, \ldots, r_k]$ and $T$. We introduce the \textit{semantic bandit} as an extension of the MAB that makes this additional structure explicit. A semantic bandit instance is defined by the standard MAB tuple, an \textit{action nomenclature}, and a \textit{scenario}. The action nomenclature is the set of textual labels $L = [l_1, \ldots, l_k] \subset V^*$ assigned to arms $[a_1, \ldots, a_k]$. These labels have no bearing on the formal reward structure, but carry semantic associations that may predispose an LLM toward or away from particular arms. The scenario $c \in C$ is the domain in which the problem is grounded. In the classical MAB, the only information relevant to an optimal policy is the history of $(action, reward)$ pairs. Any influence of semantic details on behaviour therefore constitutes a deviation from normative reward-driven decision-making.

\section{Experimental design}

\subsection{Semantic bandit specification}
\label{sec:semantic_bandit}
We instantiate our semantic bandit $ \{S,A,R,L,c\} $ as follows. We define a state space $S$ with just a single state, as is standard for MAB. We define a set of $k=3$ actions $A = [a_{H}, a_{M}, a_{L}]$, denoting the high, medium, and low reward actions respectively \footnote{For a comparison of results with 3 arms vs 5 arms, see Appendix \ref{app:arm_comparison}}. Each of these actions has an associated label in $L = [l_H,l_M,l_L]$ and an associated Gaussian reward distribution $r_i \sim \mathcal{N}(\mu_i, \sigma^2)$ with means $\mu_H,\mu_M,\mu_L$ respectively and shared variance $\sigma^2$. We select $c$ from a set of three scenarios: bandit, farm, and clothing recommendation.

We design our experiments to answer two research questions about LLM-based agents deployed on semantic bandit problems:

\begin{itemize}
    \item \textbf{RQ1}: How does action nomenclature impact exploration behaviour and
    cumulative regret?
    \item \textbf{RQ2}: How does the observed reward value impact exploration behaviour,
    and are there threshold effects consistent with an expected-scale bias?
\end{itemize}

\subsection{Experimental Conditions}

\subsubsection{Action nomenclature (RQ1)}
\label{sec:nomenclature}

We evaluate four action nomenclatures, each designed to engage a different
type of semantic prior. Table \ref{tab:nomenclatures} shows examples of each for the farming scenario. The \textit{alphanumeric} nomenclature uses randomly sampled six-character strings (e.g., \texttt{f4tjo5}) and serves as a control condition with minimal semantic content. The \textit{sentiment} nomenclature assigns adjectives of positive, neutral, and negative valence to the high-, medium-, and low-reward arms respectively, targeting the positivity bias documented in LLMs \citep{sharma2024towards}; labels are sampled at the start of each run from \citet{taboada-etal-2011-lexicon}. The \textit{ordinal} nomenclature uses explicit rank labels (\textit{Highest}, \textit{Intermediate}, \textit{Lowest}), inducing a direct semantic ordering. The \textit{world knowledge} nomenclature uses domain-specific labels whose relative value can be inferred from pre-training knowledge. We do not include a world nomenclature for the bandit scenario (see Section \ref{sec:scenarios}).

To disentangle the effect of nomenclature from the effect of reward feedback, we evaluate each nomenclature (besides alphanumeric) in both a \textit{helpful} configuration --- where the semantically favoured label is assigned to the highest-reward arm --- and a \textit{misleading} configuration, where the labels for the highest- and lowest-reward actions are swapped .

\begin{table*}[ht]
\centering
\resizebox{\linewidth}{!}{%
\renewcommand{\arraystretch}{1.2}
\begin{tabular}{|l|l|l|l|l|}
\hline
\textbf{Mean Reward} & \textbf{World Knowledge} & \textbf{Ordinal} & \textbf{Sentiment} & \textbf{Alphanumeric} \\
\hline
75 & Productive Grassland & Highest & Phenomenal & f4tjo5 \\
50 & Rocky Hills & Intermediate & Typical & 88pdak \\
25 & Arid Desert & Lowest & Repulsive & 4y11s9 \\
\hline
\end{tabular}}
\caption{Helpful action nomenclature examples for the \textit{High+} reward scale. In the misleading case, the labels for the highest and lowest rewards are swapped. Note that the labels for sentiment and alphanumeric are examples - we randomly sample these labels from a set at the start of each run to reduce the bias from a single label.}
\label{tab:nomenclatures}
\end{table*}

\subsubsection{Reward scale (RQ2.1)}
\label{sec:reward_scale}
To investigate the influence of observed reward values on exploration, our main experiments contrast two reward conditions: \textit{High+}, with mean rewards $(\mu_H, \mu_M, \mu_L) = (75, 50, 25)$, and \textit{High-}, with $(\mu_H, \mu_M, \mu_L) = (-25, -50, -75)$. We also evaluated a \textit{Low+} and \textit{Low-} nomenclature, where all values are scaled down by a factor of 100. For space, we move these results to Appendix \ref{app:results}. We use a shared variance $\sigma^2$ across all arms. We evaluate three levels of variance, $\sigma \in \{12.5, 6.25, 0\}$, referred to as high, low, and no variance, as a robustness check. 

\subsubsection{Reward scale sweep (RQ2.2)}
\label{sec:scalesweep_def}
The main reward experiment does not allow us to isolate the effects of polarity and magnitude. To isolate their effects, we run a scale sweep that varies a single observed reward value $r \in [-2, 2]$ in increments of $0.1$ and measures exploration behaviour as a function of $r$ alone. The protocol is as follows: the model's first action is forced to the semantically favoured arm, which returns the specified reward $r$. We then record the number of subsequent turns until the model selects a different arm. This is repeated for each value of $r$, allowing us to characterize both the polarity threshold (is the model more likely to explore after observing negative values?) and magnitude effects (are there non-monotonic threshold effects at particular reward values?). 

\subsection{Scenarios}
\label{sec:scenarios}

We evaluate all conditions across a set of three scenarios $C$ that vary the thematic framing of
the prompt: (i) a \textit{bandit} scenario, in which the agent is explicitly told it
is solving a multi-armed bandit problem; (ii) a \textit{farming} scenario, in which
the agent selects between agricultural fields to maximise crop yield; and (iii) a
\textit{clothing recommendation} scenario, in which the agent recommends clothing
items to users. The three scenarios are not independent of task difficulty. In the
clothing scenario, stable contextual information --- describing user preferences and climate conditions --- is provided at the start of each replicate. This context is fixed within a replicate but varies across replicates. We treat the three scenarios as a robustness check on the generality of our findings rather than as a primary experimental variable. 

\subsection{Models and baselines}
\label{sec:models}

We evaluate three instruction-tuned LLMs: (i)~OLMo-3.1-32B-Instruct, (ii)~Qwen3-32B\footnote{We ran additional experiments with other Qwen3 sizes. See Appendix \ref{app:qwen_scaling}}, and
(iii)~Gemini~3.1 Flash Lite. Qwen3 and Gemini offer native support for thinking mode, while we use CoT prompting for OLMo. We run 10 replicates per condition, resampling action labels and shuffling arm order at the start of each replicate. We compare LLM agents against a classical MAB baseline: UCB1 \citep{auer_ucb_2002} \footnote{UCB observes only scalar rewards, and so is label-agnostic. We ran 10,000 replicates per condition, with arm-to-mean assignments shuffled independently at the start of each replicate.}. We adapt our prompt structure from the best-performing prompt identified by \citet{krishna_icl}, with the interaction history summarized as the observed mean reward. Additionally, we experiment with two alternative prompting strategies intended to mitigate bias from the labels: 1) an \textbf{explicit warning} that labels may be misleading (\textit{"Be careful about bias in the action labels. Action names may be misleading and are not guaranteed to correlate with reward."}) and 2) the explicit warning plus an explicit \textbf{instruction to explore} (\textit{"Make sure to explore"}). The full prompt and the prompt variations are provided in Appendix~\ref{app:prompts}. 

\subsection{Metrics}
\label{sec:metrics}

\paragraph{Cumulative regret}
We report normalized cumulative regret, defined as the cumulative difference between
the reward of the optimal arm and the reward received, normalized to $[0, 1]$. Lower
regret indicates more effective exploration--exploitation behaviour. Let $a_t$ denote the action chosen by the agent at timestep $t$, and let $a^* = \arg\max_{a \in A} r(a)$ denote the optimal action with the highest expected reward. The instantaneous regret at timestep $t$ is defined as $\ell_t = r(a^*) - r(a_t)$. The cumulative regret over a horizon of $T$ timesteps is then $R_T = \sum_{t=1}^{T} \ell_t = \sum_{t=1}^{T} \bigl(r(a_H) - r(a_t)\bigr)$
We normalize by the maximum possible regret to get the normalized cumulative regret:
$\hat{R}_T = \frac{R_T}{\sum_{t=1}^{T} \bigl(r(a^H) - r(a^L)\bigr)}$.

\paragraph{Exploration count}
We track the number of distinct arms selected across the episode. Since $k = 3$, this
value ranges from 1 (no exploration) to 3 (full exploration). We report exploration
count per turn to characterize the trajectory of exploration behaviour over the
episode rather than just its aggregate.

\paragraph{Exploration Probability}
In the scalesweep experiments, we manually set the interaction history such that at turn 0, the semantically favoured arm was selected. Let this action be $a_0$. We define $P(exploration) = P(a_1 \neq a_0)$, i.e. the probability that the model selects a different arm at turn 1. 

\subsection{Additional implementation details}
To account for stochasticity in model outputs, we run 10 replicates of each main experiment configuration. For the the reward scale sweep, we ran 20 replicates per configuration. To reduce variance from any single label choice within a nomenclature, arm order in the prompt is shuffled and labels are resampled at the start of each replicate for sentiment and alphanumeric, as well as for world on the clothing domain.

\begin{table}[ht]
\renewcommand{\arraystretch}{0.9}
\centering
\small
\resizebox{\columnwidth}{!}{%
\begin{tabular}{lr|rr|rr|rr}
\toprule
 &  & \multicolumn{2}{c}{Bandit} & \multicolumn{2}{c}{Farm} & \multicolumn{2}{c}{Clothing} \\
\cmidrule(lr){3-4} \cmidrule(lr){5-6} \cmidrule(lr){7-8}
Model & Nomenclature & Helpful & Misleading & Helpful & Misleading & Helpful & Misleading \\
\midrule
Qwen3-32B & Ordinal & \textbf{0.00} & \textbf{0.79} & \textbf{0.03} & \textbf{0.71} & \textbf{0.01} & \textbf{0.90} \\
 & World & -- & -- & 0.10 & 0.28 & 0.04 & 0.86 \\
 & Sentiment & 0.09 & 0.29 & 0.14 & 0.29 & 0.10 & 0.34 \\
 & Alphanumeric & 0.16 & -- & 0.15 & -- & 0.16 & -- \\
\midrule
OLMo-3.1 32B & Ordinal & \textbf{0.04} & \textbf{0.79} & 0.03 & 0.91 & \textbf{0.00} & \textbf{1.00} \\
 & World & -- & -- & \textbf{0.01} & \textbf{0.97} & 0.07 & 0.91 \\
 & Sentiment & 0.28 & 0.51 & 0.43 & 0.55 & 0.10 & 0.90 \\
 & Alphanumeric & 0.29 & -- & 0.47 & -- & 0.47 & -- \\
\midrule
Gemini 3.1 Flash Lite & Ordinal & \textbf{0.13} & \textbf{0.27} & \textbf{0.15} & \textbf{0.17} & 0.15 & 0.22 \\
 & World & -- & -- & \textbf{0.15} & 0.15 & \textbf{0.07} & \textbf{0.56} \\
 & Sentiment & 0.15 & 0.15 & \textbf{0.15} & 0.15 & 0.15 & 0.15 \\
 & Alphanumeric & 0.15 & -- & \textbf{0.15} & -- & 0.15 & -- \\
\midrule
UCB1 &  & 0.15 & -- & 0.15 & -- & 0.15 & -- \\
\bottomrule
\end{tabular}%
}
\caption{Normalized cumulative regret at the final turn, by model and nomenclature type with high reward scale and no variance. We bold the lowest cumulative regret for each model in the helpful case and the highest in the misleading case. These always occur from the same nomenclature.}
\label{tab:label_channel_bias_No}
\end{table}

\section{Results}

%%%%%%%%%% Gemini and OLMO nom exploration
\begin{figure*}[ht]
  \begin{center}
    \centerline{\includegraphics[width=0.95\linewidth]{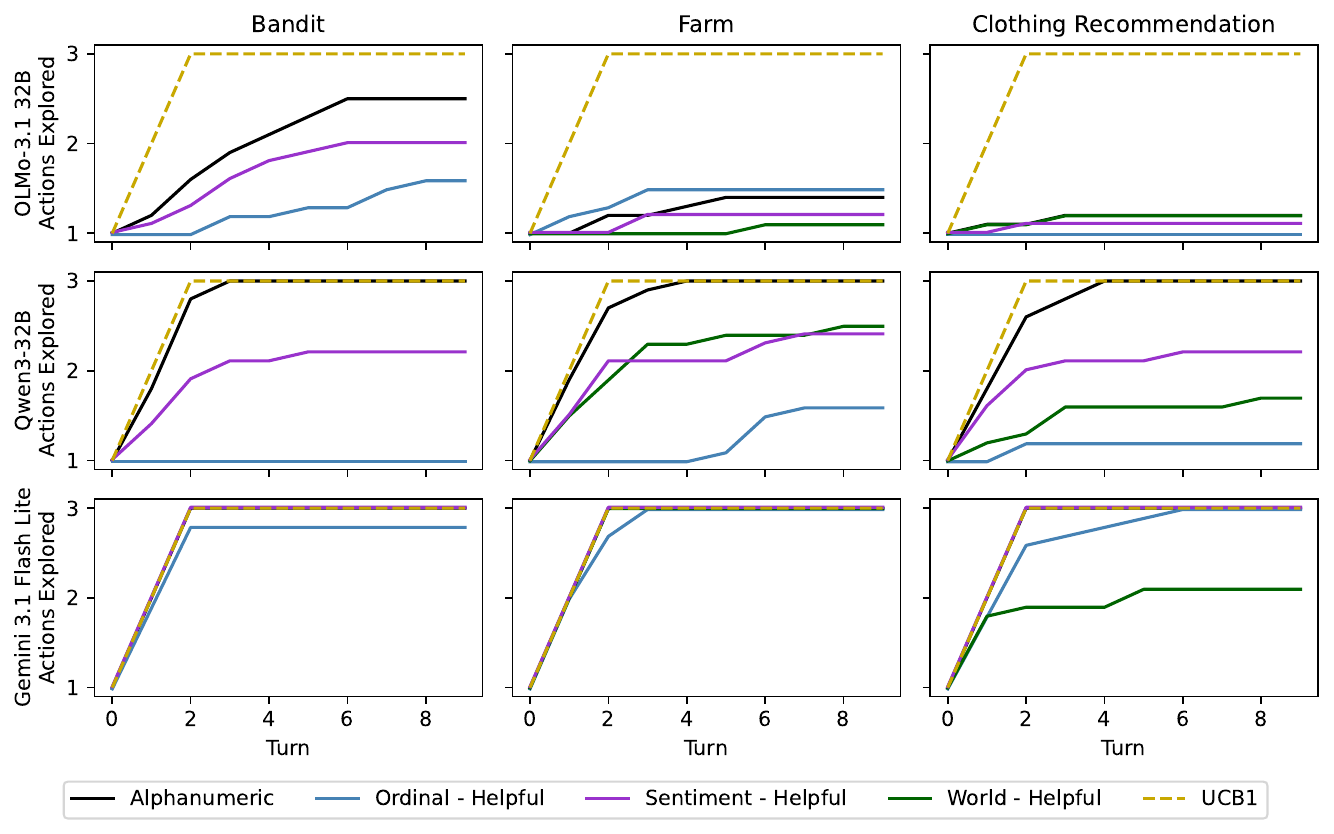}}
  \end{center}
  \caption{
      \textbf{Impact of nomenclature on exploration count.} Each row reflects a different LLM. Alphanumeric (black) tends to result in normative exploration behaviour (comparable to UCB1 baseline in yellow). In contrast, ordinal, world, and sentiment nomenclatures bias the model towards exploitation of semantically favoured actions. These results are with positive reward, helpful nomenclatures, and no variance. See full results in Appendix \ref{app:results}.
      }
    \label{fig:nom_impacts_exploration}
\end{figure*}
%%%%%%%%%% END

\subsection{RQ1 - How does action nomenclature impact performance and exploration behaviour?}
\textbf{Action nomenclatures can significantly bias the model away from normative reward-driven behaviour}. The alphanumeric nomenclature tends to result in exploration behaviour that closely matches the UCB symbolic baseline (Figure \ref{fig:nom_impacts_exploration}). In contrast, the LLM tends to exploit semantic biases when the nomenclature is semantically meaningful. Exploiting helpful semantic biases can lead to much lower regret than symbolic methods (Table \ref{tab:label_channel_bias_No}). However, this behaviour can lead to dismal performance if the semantic prior is not aligned with the underlying reward structure.

The model has a strong bias in cases where the arms are explicitly ordered (\textit{ordinal} in Figure \ref{fig:nom_impacts_exploration}). There is a weaker but still impactful bias towards words with positive sentiment. The bias from world knowledge varies, but is stronger in the clothing recommendation task than the farm task. 

Each LLM displayed distinct exploration behaviour (Figure \ref{fig:nom_impacts_exploration}). OLMo struggled to explore at all, except when explicitly informed that the task was a MAB. Qwen3 displayed both consistent exploration in the alphanumeric case and pure exploitation in the ordinal case. Gemini tended to explore quickly, matching reward-normative behaviour on most tasks. The notable exception was the world-knowledge nomenclature on the clothing recommendation task. This task was slightly more complex than the rest, since it required additional contextual knowledge in the prompt. This may indicate that the other task-nomenclature pairs are too easy for the Gemini model and it may increasingly rely on semantic priors as task complexity increases. We provide a sample reasoning trace in Appendix \ref{app:reasoning_trace} which demonstrates that while semantic priors may be reduced in large frontier reasoning models, they are still present and impact decision-making.

\begin{figure*}[ht]
  \begin{center}
    \centerline{\includegraphics[width=\linewidth]{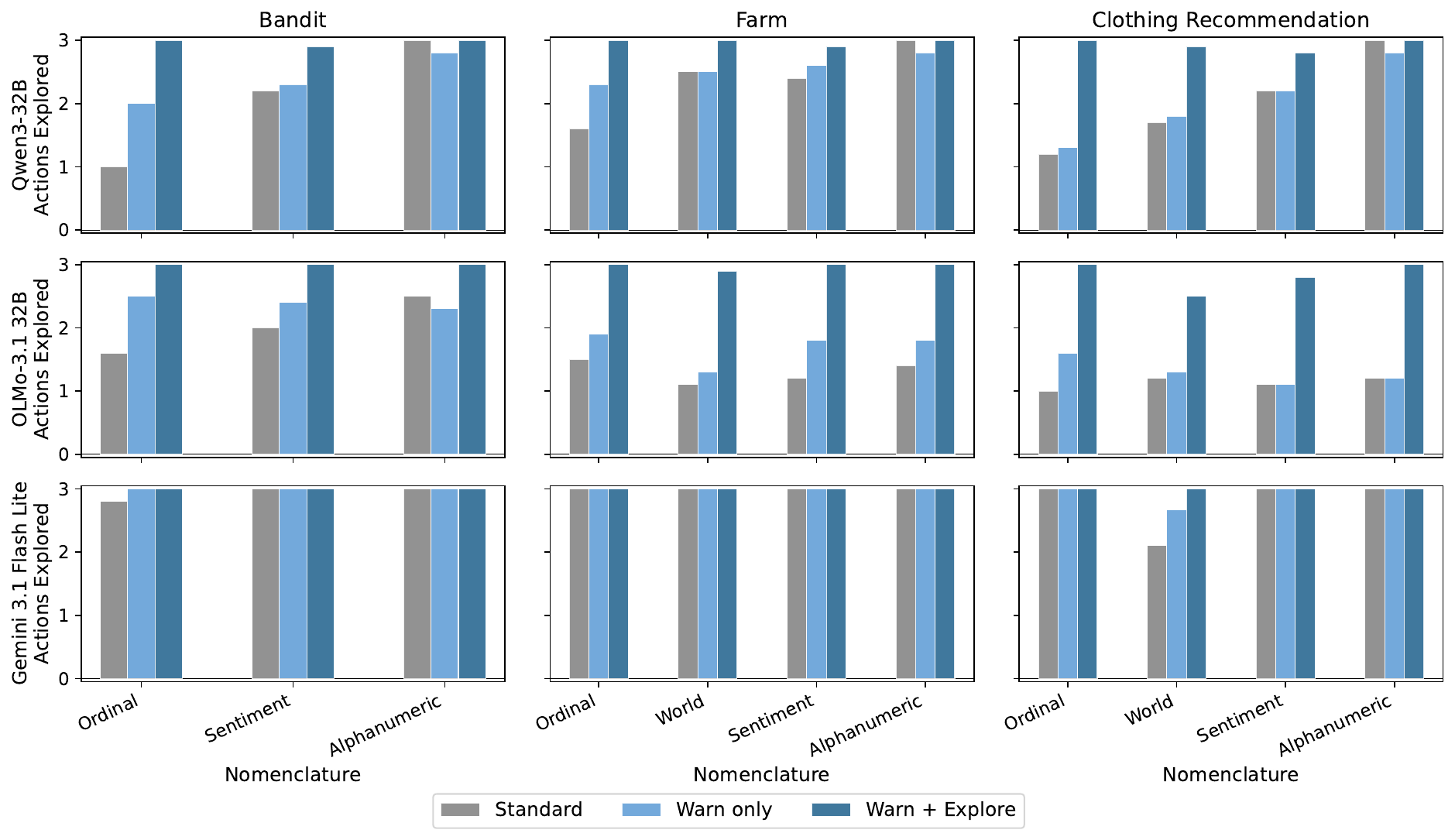}}
  \end{center}
  \caption{
      \textbf{Impact of Explicit Debiasing on Exploration Count} When the prompt includes both the warning and the instruction to explore, exploration coverage is raised significantly. When the prompt includes only the warning that biases may be misleading, exploration count tends to slightly increase. These results are with helpful nomenclatures, no variance, and high scales. See Appendix \ref{app:explicit_debiasing} for detailed results. \label{fig:explicit_debiasing_exploration_main}
    }
 \end{figure*}

\paragraph{Prompt debiasing interventions can reduce the impact of action nomenclature bias.} When the prompt includes both the warning and the instruction to explore, exploration coverage is raised significantly (Figure \ref{fig:explicit_debiasing_exploration_main}). This is unsurprising. The model no longer needs to determine how to balance between exploration and exploitation, as it has been directly instructed to explore. When the prompt includes only the warning that biases may be misleading, exploration count tends to slightly increase. The impact on cumulative regret depends on whether the nomenclature is helpful or misleading. When the nomenclature is helpful, both the instruction to explore and the warning usually increase cumulative regret. When the nomenclature is misleading, both usually decrease cumulative regret. The strength of this impact is highly inconsistent across domains, nomenclatures, and models.

\label{sec:scale_sweep}

%%%%%%%%%% Olmo and Gemini Scale Exploration
\begin{figure*}[ht]
  \begin{center}
    \centerline{\includegraphics[width=0.98\linewidth]{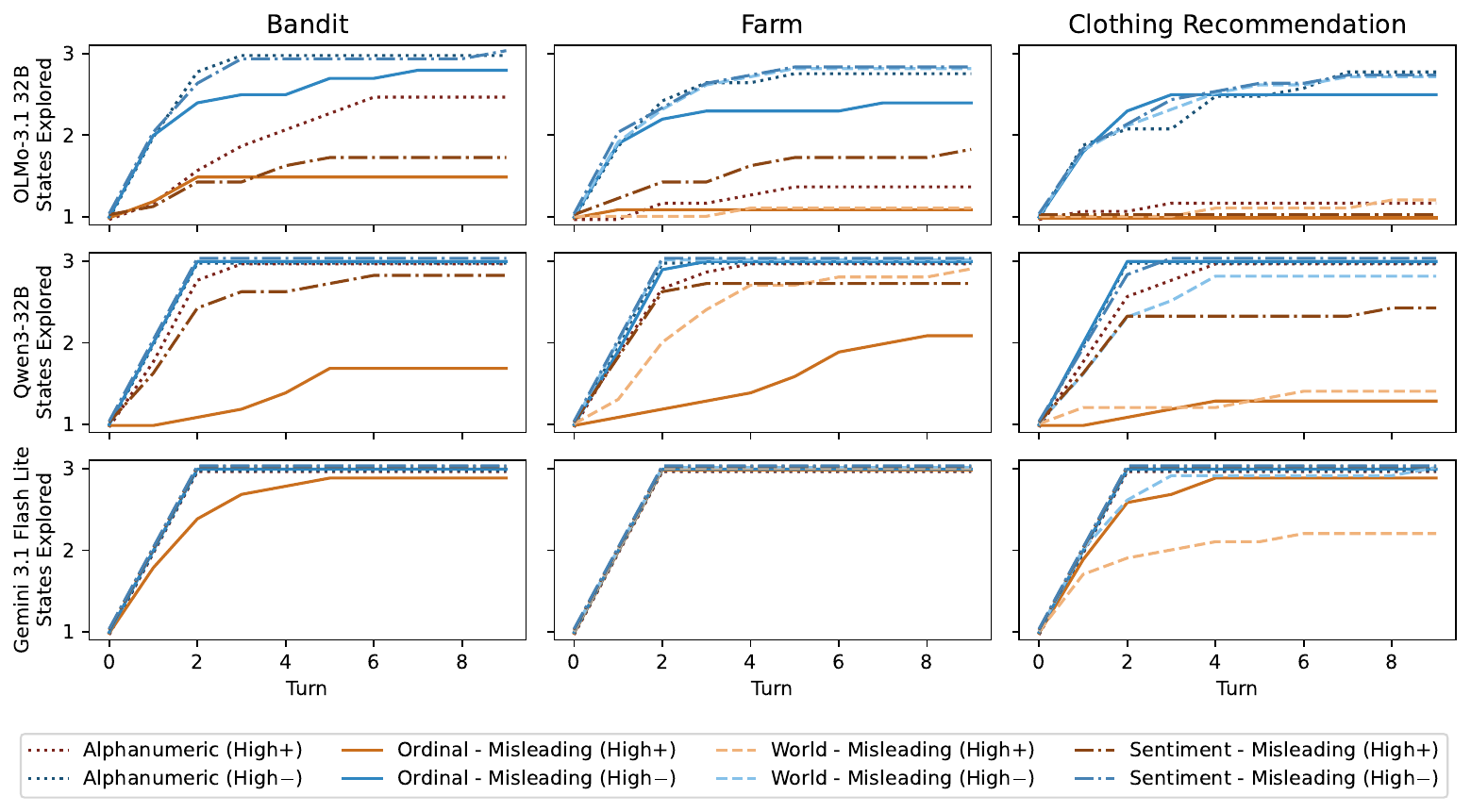}}
  \end{center}
  \caption{
      \textbf{Impact of Reward Scale on Exploration Count} We observe more exploration when reward is negative (blue) than positive (brown). These results are with no variance and \textit{High} scales. See Appendix \ref{app:results} for results with variance and with \textit{Low} scales.
    \label{fig:scale_explore}
    }
 \end{figure*}
%%%%%%%%%% END

\subsection{RQ2 - How does the observed reward value impact exploration behaviour, and are 
there threshold effects consistent with an expected-scale bias?}
Observed reward polarity is a major factor in determining model exploration behaviour. Negative rewards lead to significantly more exploration than positive rewards (blue and brown respectively in Figure \ref{fig:scale_explore} in body and Figure \ref{fig:scale_explore_low} in the appendix) The exception is when the positive reward scale has already saturated the task and reached full exploration. 

\begin{figure*}[ht]
  \begin{center}
    \subcaptionbox{Averaged across nomenclatures \label{fig:scalesweep_y0}}{
      \includegraphics[width=0.45\linewidth]{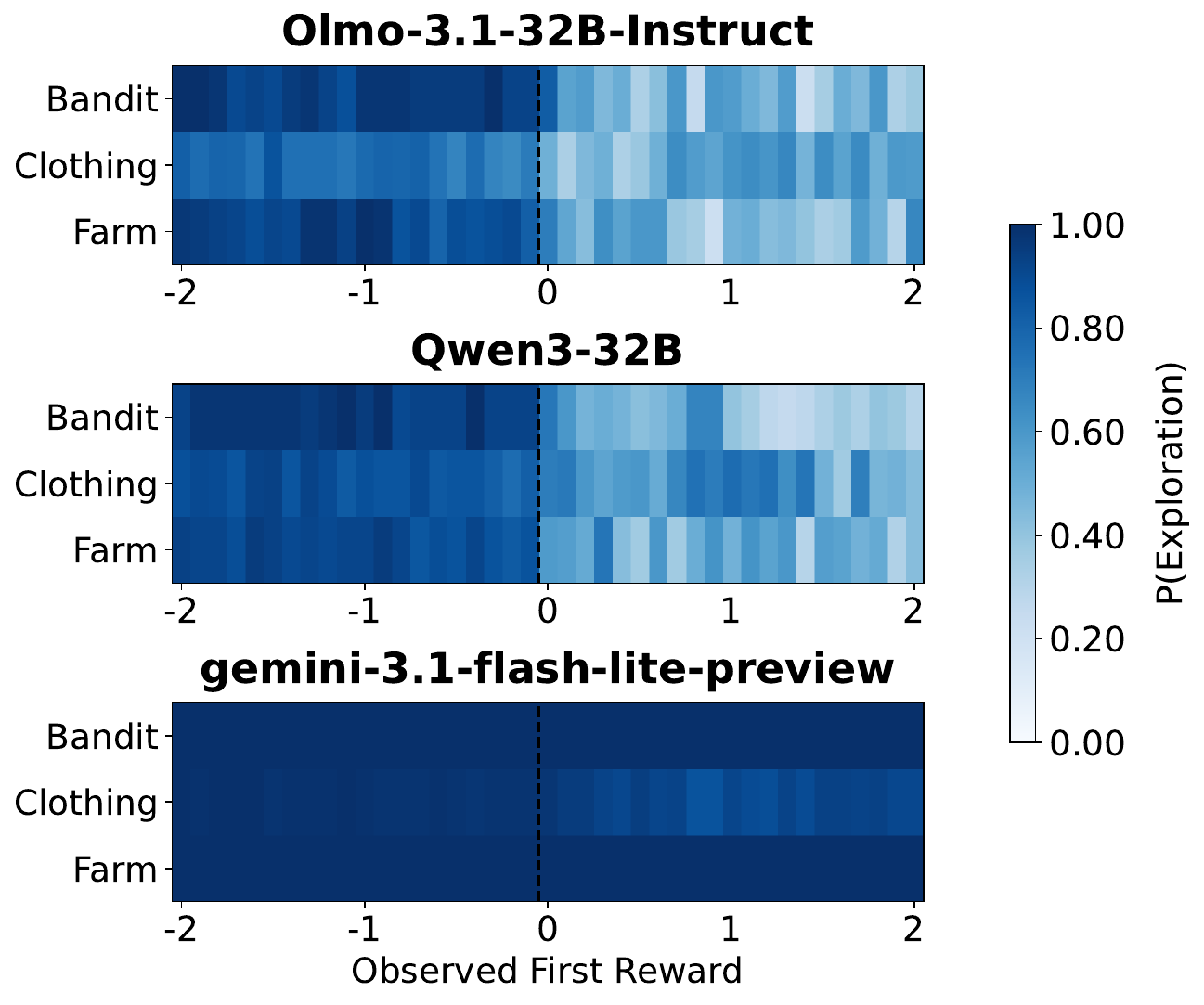}
    }
    \hfill
    \subcaptionbox{Gemini-Clothing.\label{fig:example}}{
      \includegraphics[width=0.45\linewidth]{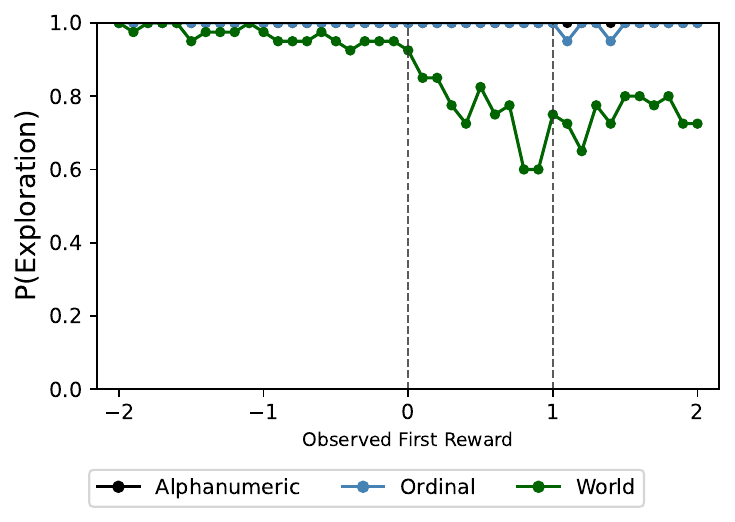}
    }
  \end{center}
  \caption{\textbf{Probability of exploration at turn 2 given the observed reward at turn 1.} We see a significantly more exploration for negative values than for positive values with OLMo and Qwen3 across all domains (left, averaged across nomenclatures). For Gemini, the effect is limited to the world nomenclature on the clothing scenario (right).}
  \label{fig:combined}
\end{figure*}

To disentangle the impact of reward sign from reward magnitude, we performed scale sweep experiments to evaluate the probability of immediate exploration given an observed reward value (Figure \ref{fig:scalesweep_y0}). We find a strong bias from reward sign --- the probability of exploring is much higher for values less than zero. This effect is larger and more sudden for Qwen3 and Olmo than for Gemini. The impact of reward magnitude is unclear and highly non-monotonic. There was weak evidence for other threshold effects around 1.0, but it was inconsistent across models and domains.  

\section{Discussion}
\textbf{LLMs are highly sensitive to surface details, sometimes overriding reward signals in a near-trivial decision-making task.} The most important finding in this paper is not that LLMs are influenced by semantic context — that is perhaps unsurprising. It is that this influence is strong enough to completely determine behaviour on one of the simplest decision-making problems that can be constructed: a stationary bandit problem with three arms and no variance. There is no noise to confuse the signal, no complexity to overwhelm the reasoner, no ambiguity about which arm is best once it has been sampled. A classical agent identifies the optimal arm within a handful of turns and never deviates. Yet changing only the textual labels assigned to arms — while holding the formal task structure constant — is sufficient to drive LLM behaviour from near-optimal to near-worst-case. While we did not explore realistic deployment contexts, failure on such a simple synthetic task raises concerns about the stability of exploration in more complex environments.

\textbf{The assumption of semantic alignment in evaluation} Our results have important implications for practitioners in environments with potentially misleading semantic biases. Practitioners who evaluate LLM agents exclusively on tasks with naturalistic, semantically coherent labels are measuring a mixture of reward-driven reasoning and semantic prior exploitation, without knowing the proportion of each. This is a concern for the generalizability of evaluation results: a model that scores well because its priors happen to align with the reward structure in the evaluation environment may fail silently in a deployment environment where that alignment does not hold.  Through prompt interventions, the model can be steered away from semantic prior exploitation and towards exploration. However, these semantic priors are often desirable and may be the reason for using an LLM over a classical solver. The practitioner must determine whether semantic priors act as a help or a hinderance in their context.

\section{Limitations and Future Work}
A major limitation of our work is the limited number of replicates we performed due to computational constraints. Further experimentation is necessary to make conclusive statistical claims. More limitations arise due to the simplicity of our semantic bandit environment. While it is effective for demonstrating the sensitivity of LLMs to semantic priors in even simple contexts, it also limits the scope of our claims. Future work could evaluate more complex bandit environments, exploring the impact of semantic priors introduced by context in contextual bandits. An important line of future work would be to ground these effects concretely in realistic agentic deployment settings. Finally, our work only considers a few avenues for semantic priors to manifest. There are a number of avenues for future work to extend this, such as the form of numeric rewards (e.g. number of decimal points) or the prompt theme itself. 

\section{Conclusion}
We introduced the semantic bandit, a formal extension of the multi-armed bandit that makes explicit the action nomenclature and scenario context through which LLM agents engage with decision tasks. We found that (1) semantically informative labels reduce exploration in favour of exploitation, improving performance under helpful alignment and severely degrading it under misalignment and (2) negative reward values trigger substantially more exploration than formally equivalent positive values, consistent with an expected-scale bias arising from pre-training conventions. Understanding when semantic context aids adaptation and when it induces systematic errors is a prerequisite for reliable deployment of LLM agents in real-world settings. We hope the semantic bandit framework provides a useful tool for future work on the robustness and calibration of LLM decision-making.

\section*{Ethics Statement}
The findings of this paper raise concerns about the deployment of LLM-based agents. We identify that semantic biases can be used to manipulate model behaviour. While we present this as a caution to practitioners, we recognize that this could also be used adversarially. Finally, we want to re-iterate that these findings correspond to a small set of experiments in a controlled environment. Practitioners should run evaluations in their own environments to determine how they are impacted.

\section*{Acknowledgments}
The authors would like to thank the reviewers and area chairs for their valuable feedback. This work was supported by the Natural Sciences and Engineering Research Council of Canada (NSERC). Jackie Chi Kit Cheung is supported by a Canada CIFAR AI Chair. We acknowledge material support from NVIDIA Corporation in the form of computational resources provided to Mila. We would like to thank Google for providing free Gemini credits as well as the Digital Research Alliance of Canada and Mila for providing additional compute.

\FloatBarrier
\bibliography{colm2026_conference}
\bibliographystyle{colm2026_conference}

\appendix
\FloatBarrier

\section{Experimental Design}
The appendix is organized into two sections. 

The first section (this one) contains experiment implementation details and experimental results associated with design choices. This includes implementation details (\ref{app:implementation}), prompt templates and nomenclature sets(\ref{app:prompts}), results using alternative prompting schemes (\ref{app:explicit_debiasing} and \ref{app:history_comparison}). 

The second section contains more granular details of the main experimental results in the paper. This includes results using the "low" reward scale and with added variance to the reward distributions. Results are stable across variance conditions. This section is broken down into cumulative regret results for main experiments (\ref{app:cum_regret}), exploration count (\ref{app:exp_count}), scaling results with different size Qwen models (\ref{app:qwen_scaling}), a comparison between a scenario with 3 arms and 5 arms (\ref{app:arm_comparison}), and sample reasoning traces (\ref{app:reasoning_trace}).

\subsection{Implementation Details}
\label{app:implementation}
To reduce computational overhead, we limit
thinking tokens to 150 for OLMo and Qwen; for Gemini, thinking tokens are
unrestricted. This allows us to contrast standard open-source models with closed-source frontier reasoning models. All other hyperparameters are set to default HuggingFace values for open-source models and Gemini standards for Gemini. 

We ran initial evaluations of Llama-3.1-8B-Instruct, Qwen3-8B, and Qwen3-14B. We found that the 8B models did not display sufficient exploration behaviour, while Qwen3-14B performed similarly to Qwen3-32B. We dropped these models in favour of the larger models due to known performance gains at scale \citep{monea_bandit}.

\subsection{Prompt Templates}
\label{app:prompts}
This section contains the prompt templates used in the main experiments.

\definecolor{jinjacolor}{HTML}{C0392B}  % red for runtime variables
\definecolor{jinjablock}{HTML}{2980B9}  % blue for conditional blocks
\definecolor{promptbg}{HTML}{F9F9F9}
\definecolor{promptframe}{HTML}{AAAAAA}

\newcommand{\jvar}[1]{{\color{jinjacolor}\texttt{\{\{#1\}\}}}}
\newcommand{\jblk}[1]{{\color{jinjablock}\texttt{\{\%\ #1\ \%\}}}}

% =====================================================================
% A.1  Prompt Template
% =====================================================================
\begin{figure}[t]
\begin{tcolorbox}[
    colback=promptbg,
    colframe=promptframe,
    title={\small\sffamily\textbf{Prompt template: \texttt{budget\_summhist\_v1}}},
    boxrule=0.4pt,
    arc=2pt,
    left=6pt, right=6pt, top=4pt, bottom=4pt,
    breakable
]
\small\ttfamily\setlength{\parindent}{0pt}\setlength{\parskip}{0.45em}
You will be assigned a task.\\
You will receive the actions you have available, followed by the action history.\\
Your goal is to maximize the total reward across all \jvar{num\_turns} steps.
We are on step \jvar{current\_turn}.

Task Assignment:\\
{-}{-}{-}\\
\jvar{task\_prompt}\\
{-}{-}{-}\\[0.1em]
\jblk{if context}\\
Current Context:\\
{-}{-}{-}\\
\hspace*{1em}- \jvar{key}: \jvar{value}\\
{-}{-}{-}\\
\jblk{endif}\\[0.1em]
Available Actions:\\
{-}{-}{-}\\
\hspace*{1em}- \jvar{action\_suffix} Name: \jvar{action} -- This action has been selected \jvar{count} time(s) for an observed mean of \jvar{obs\_mean}\\
{-}{-}{-}

When selecting an action, include only the name of the action.
Do not include the dashes used to indicate the list here.

Please select an action from the list of action options in order to maximize
the reward for your task. When selecting an action, put the exact name of the
action between action tags like:
\texttt{\textless action\textgreater{} action\_name \textless/action\textgreater}.

Your thinking budget is \jvar{thinking\_budget} tokens. If you exceed this
limit, you will be forced to provide an immediate answer.\\
Your overall generation limit is \jvar{new\_token\_budget} tokens.
\end{tcolorbox}
\caption{Prompt template used in all experiments
(\texttt{budget\_summhist\_v1}).
\textcolor{jinjacolor}{\texttt{\{\{red\}\}}} variables are filled at runtime.
The \textcolor{jinjablock}{\texttt{\{\% blue \%\}}} conditional block
(context) appears only in the clothing (contextual bandit) condition.}
\label{fig:prompt-template}
\end{figure}

% =====================================================================
% A.2  Task Prompts
% =====================================================================
\begin{table}[t]
\centering
\caption{Task prompts used for each domain.
All other template fields are held constant.}
\label{tab:task-prompts}
\small
\begin{tabularx}{\linewidth}{@{}lX@{}}
\toprule
\textbf{Domain} & \textbf{Task prompt} \\
\midrule
Farming &
You are a farmer who owns 3 fields. Each action corresponds to planting
your crop in a different field. Each field is associated with a Gaussian
distribution with a fixed but unknown mean. At each time step, you will
select a field and receive a reward drawn from the associated distribution.
Your goal is to maximize your reward across all of your turns. \\[0.6em]
Clothing &
You are a recommendation engine that chooses clothing items to display to
users when they visit your webpage. You will be presented with 3 items and a
user context. Selecting an item corresponds to recommending it to the user.
Each item is associated with a Gaussian distribution with a fixed but unknown
mean. At each time step, you will select an item and receive a reward drawn
from the associated distribution. Your goal is to maximize your reward across
all of your turns. \\[0.6em]
Abstract &
You are faced with a 3-armed bandit problem. Each action corresponds to
pulling a specific arm. Each arm is associated with a Gaussian distribution
with a fixed but unknown mean. At each time step, you will select an arm and
receive a reward drawn from the associated distribution. Your goal is to
select which arm to pull in order to maximize your reward across all of your
turns. \\
\bottomrule
\end{tabularx}
\end{table}

% =====================================================================
% A.3  Nomenclature Sets
% =====================================================================
\begin{table}[t]
\centering
\caption{Action name sets across nomenclature conditions. Each condition
assigns a name drawn from the \emph{high}, \emph{mid}, and \emph{low} pools
to the high-, mid-, and low-reward arms respectively. In the
\emph{helpful} condition this assignment is congruent with arm value; in the
\emph{mislead} condition the high and low pools are swapped. Sentiment pools
contain 20 words each; three representative examples are shown.
World (Clothing) names are context-dependent: the same item is high-reward
in a matching weather context and low-reward in a mismatched one
(representative examples from the shirt and jacket categories shown).}
\label{tab:nomenclatures}
\small
\setlength{\tabcolsep}{7pt}
\begin{tabular}{@{}llll@{}}
\toprule
\textbf{Type} & \textbf{High pool} & \textbf{Mid pool} & \textbf{Low pool} \\
\midrule
Ordinal
  & ``Highest Reward''
  & ``Intermediate Reward''
  & ``Lowest Reward'' \\[0.4em]
Sentiment
  & \textit{excellent, fabulous,}
  & \textit{standard, typical,}
  & \textit{awful, terrible,} \\
  & \textit{outstanding}
  & \textit{functional}
  & \textit{deplorable} \\[0.4em]
World (Farm)
  & Productive Grassland
  & Rocky Hills
  & Arid Desert \\[0.4em]
\multirow{3}{*}{\parbox{2.5cm}{World (Clothing)\\warm context}}
  & tank top,
  & midweight t-shirt,
  & flannel long-sleeve, \\
  & windbreaker
  & leather jacket
  & heavy parka \\[0.4em]
\multirow{3}{*}{\parbox{2.5cm}{World (Clothing)\\cold context}}
  & flannel long-sleeve,
  & midweight t-shirt,
  & tank top, \\
  & heavy parka
  & leather jacket
  & windbreaker \\
\bottomrule
\end{tabular}
\end{table}

\FloatBarrier

\subsection{Alternative History Passing Formats}
\label{app:history_comparison}
\begin{table}[ht]
\centering
\small
\resizebox{\linewidth}{!}{%
\begin{tabular}{ll|rrrr|rrrr}
\toprule
History & Nomenclature & Bandit Regret H & Bandit Regret M & Bandit Explor H & Bandit Explor M & Farm Regret H & Farm Regret M & Farm Explor H & Farm Explor M \\
\midrule
Full & Ordinal & 0.01 & 1.00 & 1.20 & 1.00 & 0.01 & 0.95 & 1.10 & 1.10 \\
 & World & -- & -- & -- & -- & 0.07 & 0.82 & 1.80 & 1.40 \\
 & Sentiment & 0.27 & 0.62 & 1.30 & 1.30 & 0.29 & 0.35 & 2.10 & 2.40 \\
 & Alphanumeric & 0.23 & -- & 2.20 & -- & 0.38 & -- & 2.00 & -- \\
\midrule
Summarized & Ordinal & 0.00 & 0.79 & 1.00 & 1.70 & 0.03 & 0.71 & 1.60 & 2.10 \\
 & World & -- & -- & -- & -- & 0.10 & 0.28 & 2.50 & 2.90 \\
 & Sentiment & 0.27 & 0.35 & 1.90 & 2.30 & 0.17 & 0.33 & 2.40 & 2.40 \\
 & Alphanumeric & 0.16 & -- & 3.00 & -- & 0.15 & -- & 3.00 & -- \\
\bottomrule
\end{tabular}%
}
\caption{Full vs. summarized history comparison at turn @9 for Qwen3-32B. Summarized history is presenting the observed mean reward and number of pulls for each arm. Full history is the raw action-reward history. Overall, we found that summarized history performed better.}
\label{tab:history_comparison}
\end{table}
\FloatBarrier

\subsection{Additional Results from Prompting with Explicit Debiasing}
\label{app:explicit_debiasing}
\begin{table}[ht]
\centering
\small
\resizebox{\linewidth}{!}{%
\begin{tabular}{lccl|rrrr|rrrr|rrrr}
\toprule
 &  &  &  & \multicolumn{4}{c|}{Bandit} & \multicolumn{4}{c|}{Farm} & \multicolumn{4}{c}{Clothing} \\
\cmidrule(lr){5-8} \cmidrule(lr){9-12} \cmidrule(lr){13-16}
 &  &  &  & \multicolumn{2}{c}{Helpful} & \multicolumn{2}{c|}{Misleading} & \multicolumn{2}{c}{Helpful} & \multicolumn{2}{c|}{Misleading} & \multicolumn{2}{c}{Helpful} & \multicolumn{2}{c}{Misleading} \\
\cmidrule(lr){5-6} \cmidrule(lr){7-8} \cmidrule(lr){9-10} \cmidrule(lr){11-12} \cmidrule(lr){13-14} \cmidrule(lr){15-16}
Model & Warn. & Instr. & Nomenclature & @2 & @9 & @2 & @9 & @2 & @9 & @2 & @9 & @2 & @9 & @2 & @9 \\
\midrule
Qwen3-32B & -- & -- & Ordinal & 1.00 & 1.00 & 1.10 & 1.70 & 1.00 & 1.60 & 1.20 & 2.10 & 1.20 & 1.20 & 1.10 & 1.30 \\
 & \checkmark & -- &  & 1.50 & 2.00 & 2.10 & 2.40 & 1.60 & 2.30 & 2.00 & 2.80 & 1.20 & 1.30 & 1.40 & 1.60 \\
 & \checkmark & \checkmark &  & 2.50 & 3.00 & 2.60 & 3.00 & 2.70 & 3.00 & 2.90 & 3.00 & 2.80 & 3.00 & 2.80 & 3.00 \\
\cmidrule(lr){4-16}
 & -- & -- & World & -- & -- & -- & -- & 1.90 & 2.50 & 2.00 & 2.90 & 1.30 & 1.70 & 1.20 & 1.40 \\
 & \checkmark & -- &  & -- & -- & -- & -- & 1.60 & 2.50 & 1.90 & 2.60 & 1.50 & 1.80 & 1.20 & 1.80 \\
 & \checkmark & \checkmark &  & -- & -- & -- & -- & 2.80 & 3.00 & 2.40 & 3.00 & 2.40 & 2.90 & 2.10 & 2.50 \\
\cmidrule(lr){4-16}
 & -- & -- & Sentiment & 1.90 & 2.20 & 2.40 & 2.80 & 2.10 & 2.40 & 2.60 & 2.70 & 2.00 & 2.20 & 2.30 & 2.40 \\
 & \checkmark & -- &  & 1.90 & 2.30 & 2.00 & 2.40 & 2.20 & 2.60 & 2.20 & 2.80 & 2.10 & 2.20 & 2.30 & 2.60 \\
 & \checkmark & \checkmark &  & 2.80 & 2.90 & 2.50 & 2.90 & 2.80 & 2.90 & 2.80 & 3.00 & 2.40 & 2.80 & 2.70 & 2.80 \\
\cmidrule(lr){4-16}
 & -- & -- & Alphanumeric & 2.80 & 3.00 & -- & -- & 2.70 & 3.00 & -- & -- & 2.60 & 3.00 & -- & -- \\
 & \checkmark & -- &  & 2.70 & 2.80 & -- & -- & 2.40 & 2.80 & -- & -- & 2.20 & 2.80 & -- & -- \\
 & \checkmark & \checkmark &  & 2.90 & 3.00 & -- & -- & 2.80 & 3.00 & -- & -- & 3.00 & 3.00 & -- & -- \\
\midrule
OLMo-3.1 32B & -- & -- & Ordinal & 1.00 & 1.60 & 1.50 & 1.50 & 1.30 & 1.50 & 1.10 & 1.10 & 1.00 & 1.00 & 1.00 & 1.00 \\
 & \checkmark & -- &  & 1.70 & 2.50 & 1.50 & 2.00 & 1.60 & 1.90 & 1.70 & 2.30 & 1.50 & 1.60 & 1.20 & 1.20 \\
 & \checkmark & \checkmark &  & 2.90 & 3.00 & 3.00 & 3.00 & 2.50 & 3.00 & 2.60 & 3.00 & 2.50 & 3.00 & 2.50 & 2.90 \\
\cmidrule(lr){4-16}
 & -- & -- & World & -- & -- & -- & -- & 1.00 & 1.10 & 1.00 & 1.10 & 1.10 & 1.20 & 1.00 & 1.20 \\
 & \checkmark & -- &  & -- & -- & -- & -- & 1.10 & 1.30 & 1.10 & 1.50 & 1.00 & 1.30 & 1.10 & 1.10 \\
 & \checkmark & \checkmark &  & -- & -- & -- & -- & 2.40 & 2.90 & 2.50 & 3.00 & 1.60 & 2.50 & 2.00 & 2.60 \\
\cmidrule(lr){4-16}
 & -- & -- & Sentiment & 1.30 & 2.00 & 1.40 & 1.70 & 1.00 & 1.20 & 1.40 & 1.80 & 1.10 & 1.10 & 1.00 & 1.00 \\
 & \checkmark & -- &  & 1.60 & 2.40 & 1.30 & 2.10 & 1.50 & 1.80 & 1.30 & 1.80 & 1.10 & 1.10 & 1.10 & 1.20 \\
 & \checkmark & \checkmark &  & 3.00 & 3.00 & 2.90 & 3.00 & 2.70 & 3.00 & 2.60 & 3.00 & 2.60 & 2.80 & 2.10 & 3.00 \\
\cmidrule(lr){4-16}
 & -- & -- & Alphanumeric & 1.60 & 2.50 & -- & -- & 1.20 & 1.40 & -- & -- & 1.10 & 1.20 & -- & -- \\
 & \checkmark & -- &  & 1.90 & 2.30 & -- & -- & 1.40 & 1.80 & -- & -- & 1.00 & 1.20 & -- & -- \\
 & \checkmark & \checkmark &  & 2.90 & 3.00 & -- & -- & 2.90 & 3.00 & -- & -- & 2.10 & 3.00 & -- & -- \\
\midrule
Gemini 3.1 Flash Lite & -- & -- & Ordinal & 2.80 & 2.80 & 2.40 & 2.90 & 2.70 & 3.00 & 3.00 & 3.00 & 2.60 & 3.00 & 2.60 & 2.90 \\
 & \checkmark & -- &  & 2.90 & 3.00 & 3.00 & 3.00 & 2.90 & 3.00 & 3.00 & 3.00 & 3.00 & 3.00 & 3.00 & 3.00 \\
 & \checkmark & \checkmark &  & 3.00 & 3.00 & 3.00 & 3.00 & 3.00 & 3.00 & 3.00 & 3.00 & 3.00 & 3.00 & 3.00 & 3.00 \\
\cmidrule(lr){4-16}
 & -- & -- & World & -- & -- & -- & -- & 3.00 & 3.00 & 3.00 & 3.00 & 1.90 & 2.10 & 1.90 & 2.20 \\
 & \checkmark & -- &  & -- & -- & -- & -- & 3.00 & 3.00 & 3.00 & 3.00 & 2.56 & 2.67 & 2.80 & 3.00 \\
 & \checkmark & \checkmark &  & -- & -- & -- & -- & 3.00 & 3.00 & 3.00 & 3.00 & 2.90 & 3.00 & 3.00 & 3.00 \\
\cmidrule(lr){4-16}
 & -- & -- & Sentiment & 3.00 & 3.00 & 3.00 & 3.00 & 3.00 & 3.00 & 3.00 & 3.00 & 3.00 & 3.00 & 3.00 & 3.00 \\
 & \checkmark & -- &  & 3.00 & 3.00 & 3.00 & 3.00 & 3.00 & 3.00 & 3.00 & 3.00 & 3.00 & 3.00 & 3.00 & 3.00 \\
 & \checkmark & \checkmark &  & 3.00 & 3.00 & 3.00 & 3.00 & 3.00 & 3.00 & 3.00 & 3.00 & 3.00 & 3.00 & 3.00 & 3.00 \\
\cmidrule(lr){4-16}
 & -- & -- & Alphanumeric & 3.00 & 3.00 & -- & -- & 3.00 & 3.00 & -- & -- & 3.00 & 3.00 & -- & -- \\
 & \checkmark & -- &  & 3.00 & 3.00 & -- & -- & 3.00 & 3.00 & -- & -- & 3.00 & 3.00 & -- & -- \\
 & \checkmark & \checkmark &  & 3.00 & 3.00 & -- & -- & 3.00 & 3.00 & -- & -- & 3.00 & 3.00 & -- & -- \\
\bottomrule
\end{tabular}%
}
\caption{Exploration count (number of distinct arms tried) at turns @2 and @9, by model, nomenclature, and debiasing condition. Warn.\,=\,explicit bias warning; Instr.\,=\,explicit exploration instruction. H\,=\,helpful framing; M\,=\,misleading framing. Scale: \texttt{high\_scale}; variance: No.}
\label{tab:comparison_label_channel_exploration}
\end{table}
\begin{table}[ht]
\centering
\small
\resizebox{\linewidth}{!}{%
\begin{tabular}{lccl|rrrr|rrrr|rrrr}
\toprule
 &  &  &  & \multicolumn{4}{c|}{Bandit} & \multicolumn{4}{c|}{Farm} & \multicolumn{4}{c}{Clothing} \\
\cmidrule(lr){5-8} \cmidrule(lr){9-12} \cmidrule(lr){13-16}
 &  &  &  & \multicolumn{2}{c}{Helpful} & \multicolumn{2}{c|}{Misleading} & \multicolumn{2}{c}{Helpful} & \multicolumn{2}{c|}{Misleading} & \multicolumn{2}{c}{Helpful} & \multicolumn{2}{c}{Misleading} \\
\cmidrule(lr){5-6} \cmidrule(lr){7-8} \cmidrule(lr){9-10} \cmidrule(lr){11-12} \cmidrule(lr){13-14} \cmidrule(lr){15-16}
Model & Warn. & Instr. & Nomenclature & @2 & @9 & @2 & @9 & @2 & @9 & @2 & @9 & @2 & @9 & @2 & @9 \\
\midrule
Qwen3-32B & -- & -- & Ordinal & 0.00 & 0.00 & 147.50 & 395.00 & 0.00 & 15.00 & 142.50 & 357.50 & 5.00 & 5.00 & 147.50 & 447.50 \\
 & \checkmark & -- &  & 17.50 & 32.50 & 112.50 & 235.00 & 22.50 & 45.00 & 117.50 & 180.00 & 12.50 & 15.00 & 132.50 & 390.00 \\
 & \checkmark & \checkmark &  & 55.00 & 82.50 & 92.50 & 122.50 & 65.00 & 105.00 & 80.00 & 120.00 & 67.50 & 82.50 & 85.00 & 95.00 \\
\cmidrule(lr){4-16}
 & -- & -- & World & -- & -- & -- & -- & 32.50 & 50.00 & 107.50 & 137.50 & 7.50 & 22.50 & 140.00 & 432.50 \\
 & \checkmark & -- &  & -- & -- & -- & -- & 25.00 & 55.00 & 115.00 & 225.00 & 20.00 & 27.50 & 145.00 & 395.00 \\
 & \checkmark & \checkmark &  & -- & -- & -- & -- & 65.00 & 87.50 & 97.50 & 132.50 & 50.00 & 77.50 & 110.00 & 250.00 \\
\cmidrule(lr){4-16}
 & -- & -- & Sentiment & 32.50 & 42.50 & 100.00 & 147.50 & 47.50 & 70.00 & 90.00 & 145.00 & 42.50 & 47.50 & 92.50 & 170.00 \\
 & \checkmark & -- &  & 40.00 & 70.00 & 80.00 & 145.00 & 52.50 & 67.50 & 95.00 & 150.00 & 40.00 & 42.50 & 100.00 & 190.00 \\
 & \checkmark & \checkmark &  & 67.50 & 97.50 & 87.50 & 122.50 & 70.00 & 107.50 & 82.50 & 92.50 & 75.00 & 92.50 & 85.00 & 122.50 \\
\cmidrule(lr){4-16}
 & -- & -- & Alphanumeric & 75.00 & 80.00 & -- & -- & 70.00 & 77.50 & -- & -- & 67.50 & 80.00 & -- & -- \\
 & \checkmark & -- &  & 80.00 & 117.50 & -- & -- & 62.50 & 72.50 & -- & -- & 70.00 & 102.50 & -- & -- \\
 & \checkmark & \checkmark &  & 72.50 & 77.50 & -- & -- & 75.00 & 97.50 & -- & -- & 75.00 & 80.00 & -- & -- \\
\midrule
OLMo-3.1 32B & -- & -- & Ordinal & 0.00 & 20.00 & 132.50 & 395.00 & 7.50 & 15.00 & 140.00 & 455.00 & 0.00 & 0.00 & 150.00 & 500.00 \\
 & \checkmark & -- &  & 45.00 & 72.50 & 125.00 & 315.00 & 35.00 & 50.00 & 92.50 & 215.00 & 47.50 & 85.00 & 135.00 & 432.50 \\
 & \checkmark & \checkmark &  & 72.50 & 82.50 & 75.00 & 85.00 & 52.50 & 90.00 & 87.50 & 95.00 & 60.00 & 80.00 & 90.00 & 125.00 \\
\cmidrule(lr){4-16}
 & -- & -- & World & -- & -- & -- & -- & 0.00 & 5.00 & 150.00 & 485.00 & 17.50 & 35.00 & 142.50 & 455.00 \\
 & \checkmark & -- &  & -- & -- & -- & -- & 2.50 & 7.50 & 137.50 & 420.00 & 15.00 & 45.00 & 137.50 & 452.50 \\
 & \checkmark & \checkmark &  & -- & -- & -- & -- & 50.00 & 70.00 & 90.00 & 95.00 & 22.50 & 60.00 & 107.50 & 205.00 \\
\cmidrule(lr){4-16}
 & -- & -- & Sentiment & 72.50 & 140.00 & 82.50 & 255.00 & 75.00 & 215.00 & 92.50 & 275.00 & 17.50 & 52.50 & 135.00 & 450.00 \\
 & \checkmark & -- &  & 72.50 & 147.50 & 82.50 & 222.50 & 60.00 & 167.50 & 92.50 & 260.00 & 62.50 & 185.00 & 117.50 & 362.50 \\
 & \checkmark & \checkmark &  & 75.00 & 85.00 & 72.50 & 90.00 & 70.00 & 82.50 & 72.50 & 92.50 & 80.00 & 120.00 & 92.50 & 127.50 \\
\cmidrule(lr){4-16}
 & -- & -- & Alphanumeric & 75.00 & 145.00 & -- & -- & 75.00 & 237.50 & -- & -- & 80.00 & 237.50 & -- & -- \\
 & \checkmark & -- &  & 77.50 & 157.50 & -- & -- & 85.00 & 220.00 & -- & -- & 90.00 & 285.00 & -- & -- \\
 & \checkmark & \checkmark &  & 77.50 & 87.50 & -- & -- & 72.50 & 77.50 & -- & -- & 85.00 & 107.50 & -- & -- \\
\midrule
Gemini 3.1 Flash Lite & -- & -- & Ordinal & 65.00 & 65.00 & 95.00 & 132.50 & 60.00 & 75.00 & 75.00 & 82.50 & 60.00 & 75.00 & 87.50 & 110.00 \\
 & \checkmark & -- &  & 70.00 & 75.00 & 75.00 & 83.33 & 70.00 & 77.50 & 75.00 & 75.00 & 75.00 & 75.00 & 75.00 & 77.50 \\
 & \checkmark & \checkmark &  & 75.00 & 82.50 & 75.00 & 115.00 & 75.00 & 87.50 & 75.00 & 95.00 & 75.00 & 90.00 & 75.00 & 115.00 \\
\cmidrule(lr){4-16}
 & -- & -- & World & -- & -- & -- & -- & 75.00 & 75.00 & 75.00 & 75.00 & 25.00 & 35.00 & 110.00 & 277.50 \\
 & \checkmark & -- &  & -- & -- & -- & -- & 75.00 & 75.00 & 75.00 & 75.00 & 52.78 & 58.33 & 80.00 & 90.00 \\
 & \checkmark & \checkmark &  & -- & -- & -- & -- & 75.00 & 80.00 & 75.00 & 91.67 & 70.00 & 80.00 & 75.00 & 125.00 \\
\cmidrule(lr){4-16}
 & -- & -- & Sentiment & 75.00 & 75.00 & 75.00 & 75.00 & 75.00 & 75.00 & 75.00 & 75.00 & 75.00 & 75.00 & 75.00 & 75.00 \\
 & \checkmark & -- &  & 75.00 & 75.00 & 75.00 & 75.00 & 75.00 & 75.00 & 75.00 & 75.00 & 75.00 & 75.00 & 75.00 & 75.00 \\
 & \checkmark & \checkmark &  & 75.00 & 90.00 & 75.00 & 85.00 & 75.00 & 90.00 & 75.00 & 100.00 & 75.00 & 90.00 & 75.00 & 97.50 \\
\cmidrule(lr){4-16}
 & -- & -- & Alphanumeric & 75.00 & 75.00 & -- & -- & 75.00 & 75.00 & -- & -- & 75.00 & 75.00 & -- & -- \\
 & \checkmark & -- &  & 75.00 & 75.00 & -- & -- & 75.00 & 75.00 & -- & -- & 75.00 & 75.00 & -- & -- \\
 & \checkmark & \checkmark &  & 75.00 & 80.56 & -- & -- & 75.00 & 87.50 & -- & -- & 75.00 & 91.67 & -- & -- \\
\bottomrule
\end{tabular}%
}
\caption{Cumulative regret at turns @2 and @9, by model, nomenclature, and debiasing condition. Warn.\,=\,explicit bias warning; Instr.\,=\,explicit exploration instruction. H\,=\,helpful framing; M\,=\,misleading framing. Scale: \texttt{high\_scale}; variance: No.}
\label{tab:comparison_label_channel_regret}
\end{table}

\begin{figure*}[ht]
  \begin{center}
    \centerline{\includegraphics[width=\linewidth]{figures/appendix/comparison_c1_bars_exploration_helpful.pdf}}
  \end{center}
  \begin{center}
    \centerline{\includegraphics[width=\linewidth]{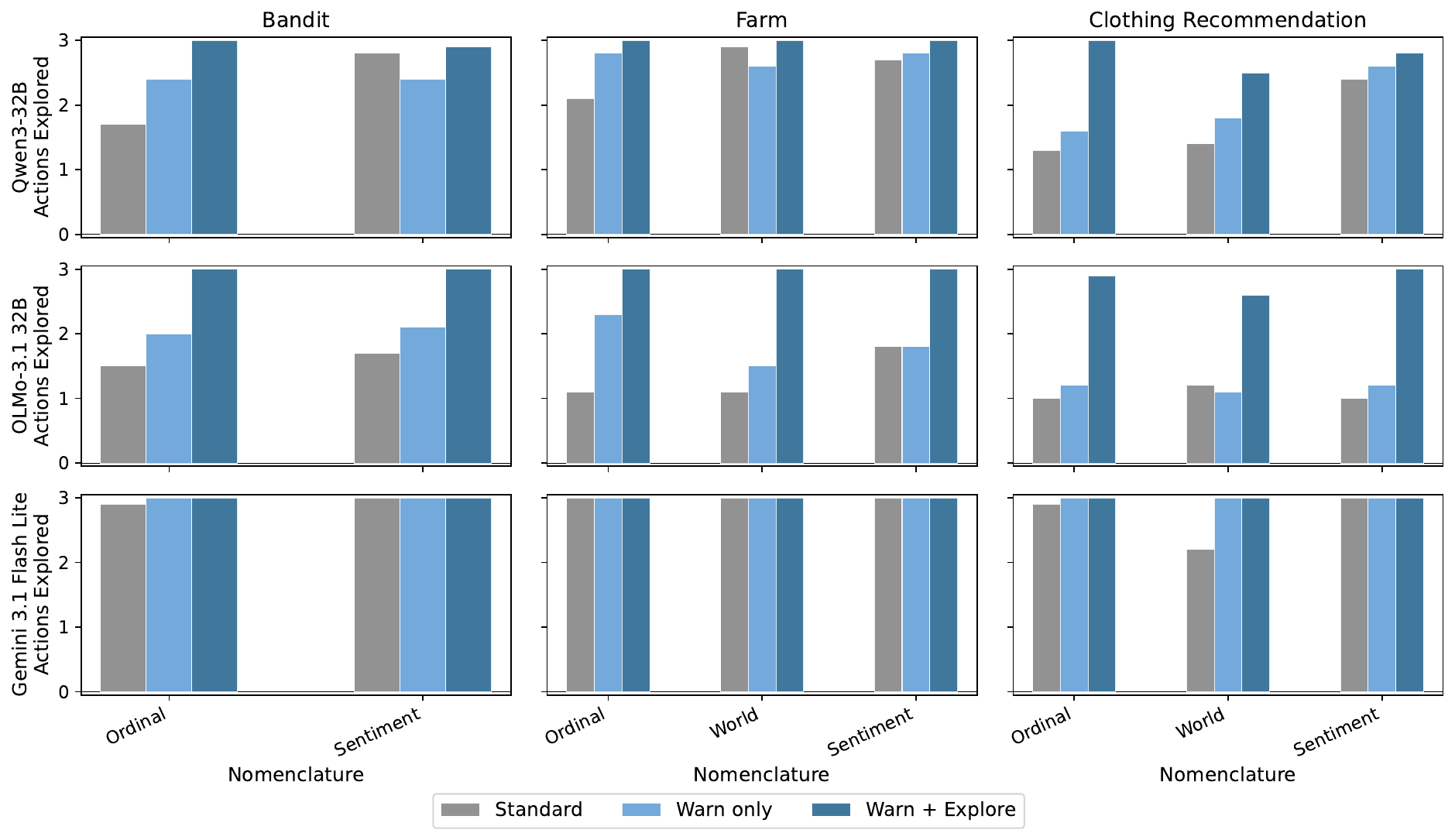}}
  \end{center}
  \caption{
      \textbf{Impact of Explicit Debiasing on Exploration Count in helpful (top) and misleading (bottom) cases.} When the prompt includes both the warning and the instruction to explore, exploration coverage is raised significantly. When the prompt includes only the warning that biases may be misleading, exploration count tends to slightly increase. These results are with no variance and \textbf{high scales}.
    \label{fig:explicit_debiasing_exploration}
    }
 \end{figure*}

 \begin{figure*}[ht]
  \begin{center}
    \centerline{\includegraphics[width=\linewidth]{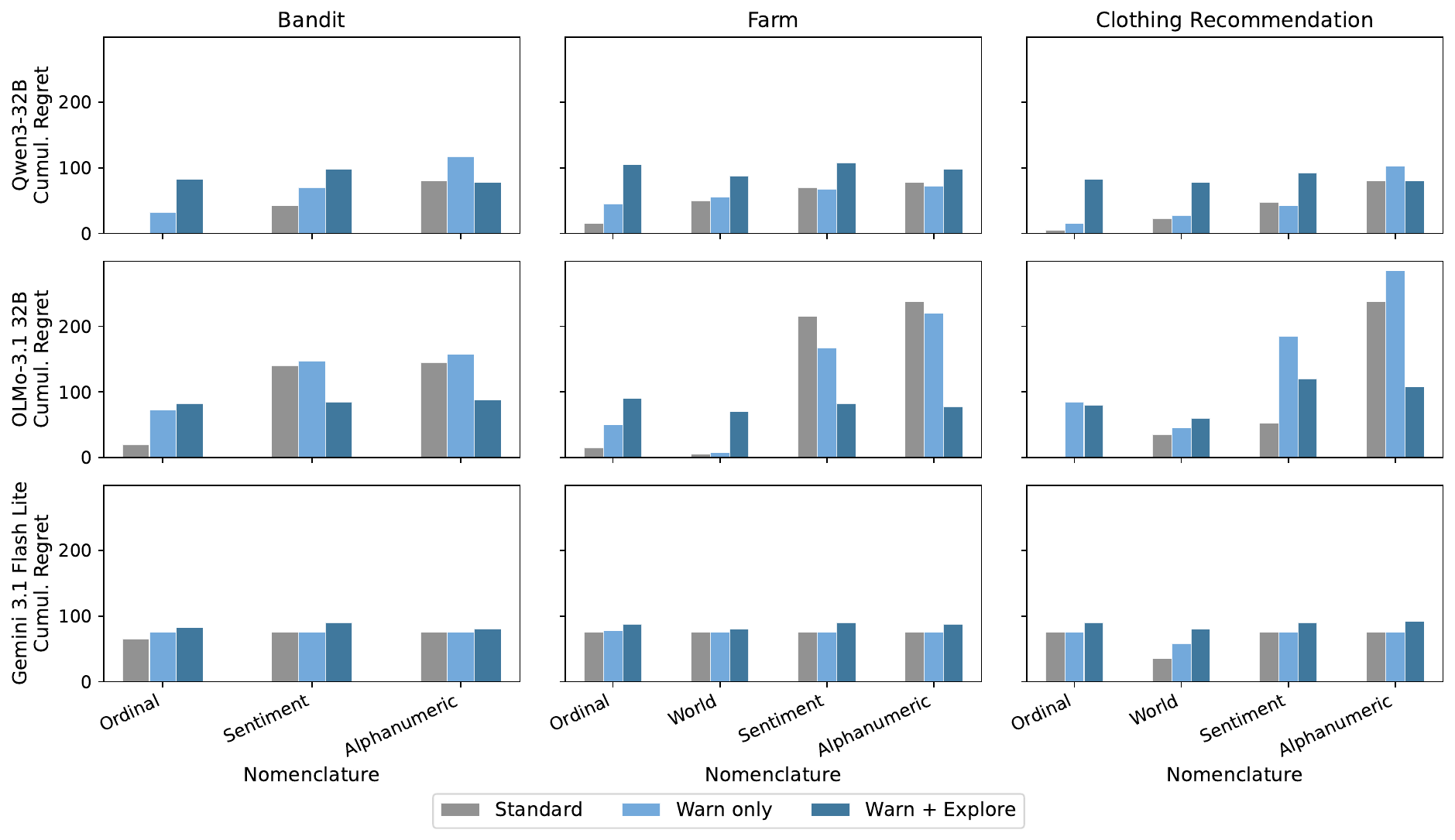}}
  \end{center}
  \begin{center}
    \centerline{\includegraphics[width=\linewidth]{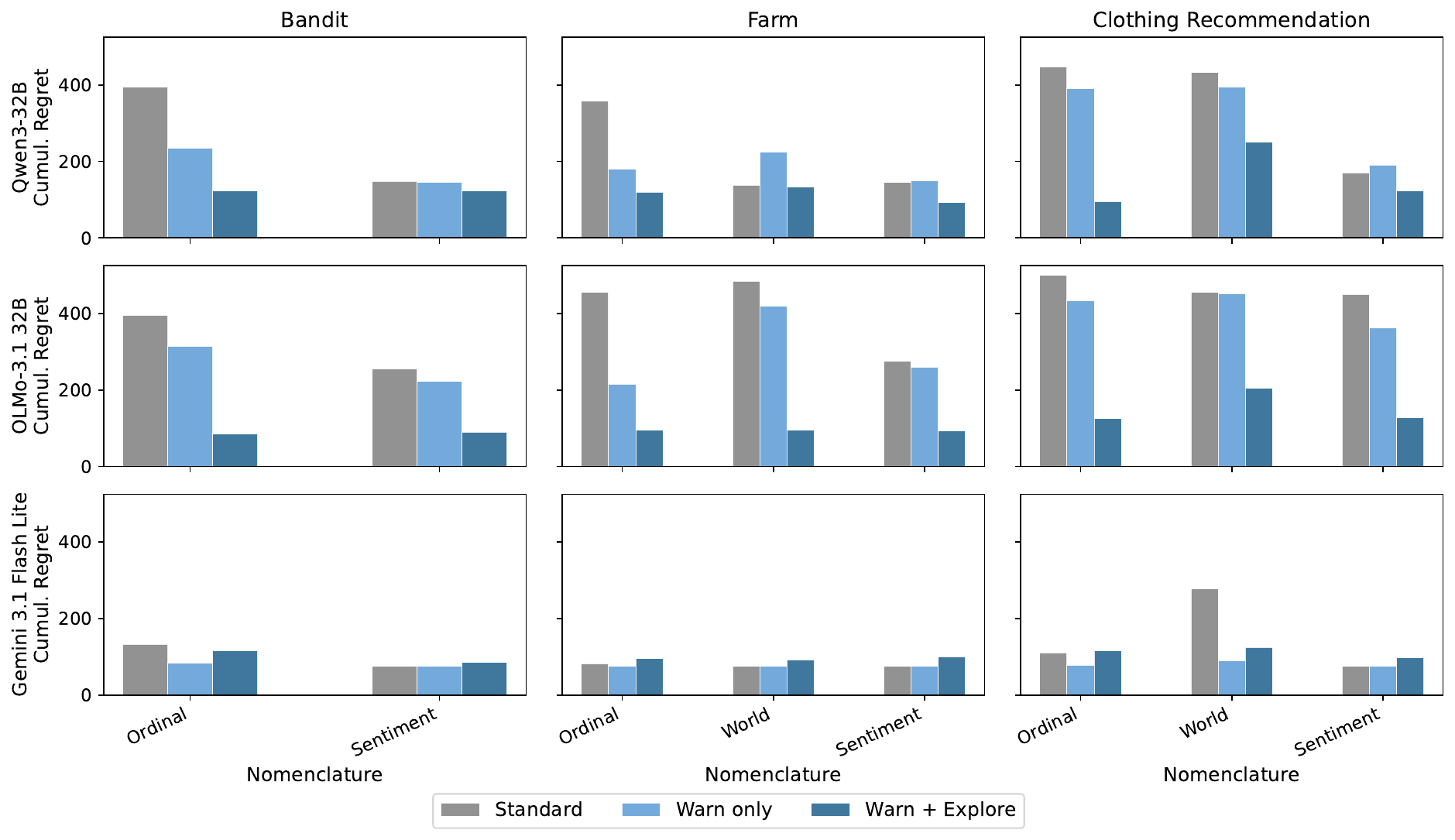}}
  \end{center}
  \caption{
      \textbf{Impact of Explicit Debiasing on Regret in helpful (top) and misleading (bottom) cases.} The impact on cumulative regret depends on whether the nomenclature is helpful or misleading. When the nomenclature is helpful, both the instruction to explore and the warning usually increase cumulative regret. When the nomenclature is misleading, both usually decrease cumulative regret. The strength of this impact is highly inconsistent across domains, nomenclatures, and models. These results are with no variance and \textbf{high scales}.
    \label{fig:explicit_debiasing_regret}
    }
 \end{figure*}

 \FloatBarrier

 \subsection{AI Disclosure}
  CoPilot and Claude Code were used in the generation of plotting code and some minor edits to core experiment functionality. 

\section{Additional Experimental Results from Main Experiments}
\label{app:results}

This section contains more detailed results for the main experiments. This includes results using the "low" reward scale and with added variance to the reward distributions. Results are stable across variance conditions. This section is broken down into cumulative regret results for main experiments (Section \ref{app:cum_regret}), exploration count (Section \ref{app:exp_count}), scaling results with different size Qwen models (\ref{app:qwen_scaling}), a comparison between a scenario with 3 arms and 5 arms (\ref{app:arm_comparison}), and sample reasoning traces (\ref{app:reasoning_trace}).

\subsection{Cumulative Regret}
\label{app:cum_regret}
These tables show the average cumulative regret for all nomenclatures, domains, models, and variance levels.
\begin{table}[H]
\centering
\small
\resizebox{\columnwidth}{!}{%
\begin{tabular}{lr|rr|rr|rr}
\toprule
 &  & \multicolumn{2}{c}{Bandit} & \multicolumn{2}{c}{Farm} & \multicolumn{2}{c}{Clothing} \\
\cmidrule(lr){3-4} \cmidrule(lr){5-6} \cmidrule(lr){7-8}
Scale & Nomenclature & Helpful & Misleading & Helpful & Misleading & Helpful & Misleading \\
\midrule
H+ & Ordinal & \textbf{0.04} & \textbf{0.79} & 0.03 & 0.91 & \textbf{0.00} & \textbf{1.00} \\
 & World & -- & -- & \textbf{0.01} & \textbf{0.97} & 0.07 & 0.91 \\
 & Sentiment & 0.28 & 0.51 & 0.43 & 0.55 & 0.10 & 0.90 \\
 & Alphanumeric & 0.29 & -- & 0.47 & -- & 0.47 & -- \\
\midrule
L+ & Ordinal & \textbf{0.08} & \textbf{0.54} & 0.03 & \textbf{0.68} & \textbf{0.00} & \textbf{0.95} \\
 & World & -- & -- & \textbf{0.03} & 0.55 & 0.07 & 0.74 \\
 & Sentiment & 0.20 & 0.35 & 0.14 & 0.30 & 0.15 & 0.73 \\
 & Alphanumeric & 0.24 & -- & 0.27 & -- & 0.30 & -- \\
\midrule
L- & Ordinal & \textbf{0.12} & \textbf{0.23} & \textbf{0.08} & \textbf{0.36} & \textbf{0.06} & \textbf{0.47} \\
 & World & -- & -- & 0.12 & 0.19 & 0.08 & 0.42 \\
 & Sentiment & 0.15 & 0.15 & 0.14 & 0.23 & 0.10 & 0.32 \\
 & Alphanumeric & 0.15 & -- & 0.20 & -- & 0.19 & -- \\
\midrule
H- & Ordinal & \textbf{0.14} & \textbf{0.28} & \textbf{0.10} & \textbf{0.42} & \textbf{0.06} & \textbf{0.38} \\
 & World & -- & -- & 0.14 & 0.27 & 0.12 & 0.38 \\
 & Sentiment & 0.16 & 0.20 & 0.14 & 0.25 & 0.10 & 0.34 \\
 & Alphanumeric & 0.16 & -- & 0.14 & -- & 0.27 & -- \\
\bottomrule
\end{tabular}%
}
\caption{Normalized cumulative regret at the final turn for OLMo-3.1 32B, variance condition: No, by reward scale and nomenclature type. Scale labels: H\,=\,high reward magnitude; L\,=\,low; +\,=\,positive rewards; $-$\,=\,negative (e.g.\ H+ is the high positive scale).}
\label{tab:regret_OLMo-31_32B_varNo}
\end{table}
\begin{table}[H]
\centering
\small
\resizebox{\columnwidth}{!}{%
\begin{tabular}{lr|rr|rr|rr}
\toprule
 &  & \multicolumn{2}{c}{Bandit} & \multicolumn{2}{c}{Farm} & \multicolumn{2}{c}{Clothing} \\
\cmidrule(lr){3-4} \cmidrule(lr){5-6} \cmidrule(lr){7-8}
Scale & Nomenclature & Helpful & Misleading & Helpful & Misleading & Helpful & Misleading \\
\midrule
H+ & Ordinal & \textbf{0.04} & \textbf{0.82} & 0.04 & \textbf{0.88} & 0.05 & 0.91 \\
 & World & -- & -- & \textbf{0.03} & 0.65 & 0.08 & \textbf{0.92} \\
 & Sentiment & 0.23 & 0.44 & 0.44 & 0.56 & \textbf{-0.00} & 0.76 \\
 & Alphanumeric & 0.27 & -- & 0.43 & -- & 0.52 & -- \\
\midrule
L+ & Ordinal & \textbf{0.07} & \textbf{0.62} & \textbf{0.03} & 0.60 & \textbf{0.00} & \textbf{0.92} \\
 & World & -- & -- & 0.05 & \textbf{0.61} & 0.05 & 0.79 \\
 & Sentiment & 0.11 & 0.39 & 0.25 & 0.42 & 0.15 & 0.64 \\
 & Alphanumeric & 0.18 & -- & 0.27 & -- & 0.32 & -- \\
\midrule
L- & Ordinal & \textbf{0.08} & \textbf{0.21} & \textbf{0.05} & \textbf{0.40} & 0.07 & 0.42 \\
 & World & -- & -- & 0.16 & 0.22 & \textbf{0.06} & \textbf{0.48} \\
 & Sentiment & 0.16 & 0.19 & 0.13 & 0.25 & 0.11 & 0.39 \\
 & Alphanumeric & 0.16 & -- & 0.20 & -- & 0.28 & -- \\
\midrule
H- & Ordinal & \textbf{0.12} & \textbf{0.26} & \textbf{0.07} & \textbf{0.43} & \textbf{0.08} & \textbf{0.47} \\
 & World & -- & -- & 0.14 & 0.23 & 0.09 & 0.33 \\
 & Sentiment & 0.16 & 0.20 & 0.17 & 0.31 & 0.14 & 0.37 \\
 & Alphanumeric & 0.14 & -- & 0.30 & -- & 0.17 & -- \\
\bottomrule
\end{tabular}%
}
\caption{Normalized cumulative regret at the final turn for OLMo-3.1 32B, variance condition: Low, by reward scale and nomenclature type. Scale labels: H\,=\,high reward magnitude; L\,=\,low; +\,=\,positive rewards; $-$\,=\,negative (e.g.\ H+ is the high positive scale).}
\label{tab:regret_OLMo-31_32B_varLow}
\end{table}
\begin{table}[H]
\centering
\small
\resizebox{\columnwidth}{!}{%
\begin{tabular}{lr|rr|rr|rr}
\toprule
 &  & \multicolumn{2}{c}{Bandit} & \multicolumn{2}{c}{Farm} & \multicolumn{2}{c}{Clothing} \\
\cmidrule(lr){3-4} \cmidrule(lr){5-6} \cmidrule(lr){7-8}
Scale & Nomenclature & Helpful & Misleading & Helpful & Misleading & Helpful & Misleading \\
\midrule
H+ & Ordinal & \textbf{0.02} & \textbf{0.89} & \textbf{0.01} & 0.94 & 0.05 & \textbf{0.94} \\
 & World & -- & -- & 0.04 & \textbf{0.98} & \textbf{-0.01} & 0.80 \\
 & Sentiment & 0.32 & 0.71 & 0.30 & 0.68 & 0.20 & 0.80 \\
 & Alphanumeric & 0.36 & -- & 0.41 & -- & 0.45 & -- \\
\midrule
L+ & Ordinal & \textbf{0.01} & \textbf{0.82} & 0.05 & 0.56 & \textbf{0.01} & \textbf{0.93} \\
 & World & -- & -- & \textbf{0.04} & \textbf{0.64} & 0.04 & 0.69 \\
 & Sentiment & 0.23 & 0.31 & 0.24 & 0.31 & 0.21 & 0.64 \\
 & Alphanumeric & 0.22 & -- & 0.29 & -- & 0.38 & -- \\
\midrule
L- & Ordinal & \textbf{0.09} & \textbf{0.36} & 0.14 & \textbf{0.38} & \textbf{0.09} & \textbf{0.44} \\
 & World & -- & -- & \textbf{0.13} & 0.17 & 0.11 & 0.40 \\
 & Sentiment & 0.15 & 0.19 & 0.17 & 0.19 & 0.16 & 0.40 \\
 & Alphanumeric & 0.16 & -- & 0.25 & -- & 0.20 & -- \\
\midrule
H- & Ordinal & \textbf{0.08} & \textbf{0.31} & \textbf{0.08} & \textbf{0.47} & \textbf{0.11} & \textbf{0.39} \\
 & World & -- & -- & 0.17 & 0.25 & 0.14 & 0.37 \\
 & Sentiment & 0.14 & 0.26 & 0.15 & 0.27 & 0.12 & 0.35 \\
 & Alphanumeric & 0.17 & -- & 0.20 & -- & 0.37 & -- \\
\bottomrule
\end{tabular}%
}
\caption{Normalized cumulative regret at the final turn for OLMo-3.1 32B, variance condition: High, by reward scale and nomenclature type. Scale labels: H\,=\,high reward magnitude; L\,=\,low; +\,=\,positive rewards; $-$\,=\,negative (e.g.\ H+ is the high positive scale).}
\label{tab:regret_OLMo-31_32B_varHigh}
\end{table}
\begin{table}[H]
\centering
\small
\resizebox{\columnwidth}{!}{%
\begin{tabular}{lr|rr|rr|rr}
\toprule
 &  & \multicolumn{2}{c}{Bandit} & \multicolumn{2}{c}{Farm} & \multicolumn{2}{c}{Clothing} \\
\cmidrule(lr){3-4} \cmidrule(lr){5-6} \cmidrule(lr){7-8}
Scale & Nomenclature & Helpful & Misleading & Helpful & Misleading & Helpful & Misleading \\
\midrule
H+ & Ordinal & \textbf{0.00} & \textbf{0.79} & \textbf{0.03} & \textbf{0.71} & \textbf{0.01} & \textbf{0.90} \\
 & World & -- & -- & 0.10 & 0.28 & 0.04 & 0.86 \\
 & Sentiment & 0.09 & 0.29 & 0.14 & 0.29 & 0.10 & 0.34 \\
 & Alphanumeric & 0.16 & -- & 0.15 & -- & 0.16 & -- \\
\midrule
L+ & Ordinal & \textbf{0.00} & \textbf{0.42} & \textbf{0.02} & \textbf{0.45} & \textbf{0.01} & 0.35 \\
 & World & -- & -- & 0.18 & 0.24 & 0.05 & \textbf{0.61} \\
 & Sentiment & 0.17 & 0.33 & 0.17 & 0.18 & 0.11 & 0.30 \\
 & Alphanumeric & 0.12 & -- & 0.15 & -- & 0.15 & -- \\
\midrule
L- & Ordinal & \textbf{0.15} & 0.16 & \textbf{0.15} & \textbf{0.20} & 0.14 & 0.16 \\
 & World & -- & -- & \textbf{0.15} & 0.18 & \textbf{0.12} & \textbf{0.48} \\
 & Sentiment & \textbf{0.15} & \textbf{0.18} & 0.18 & \textbf{0.20} & 0.16 & 0.19 \\
 & Alphanumeric & 0.15 & -- & 0.17 & -- & 0.15 & -- \\
\midrule
H- & Ordinal & \textbf{0.15} & \textbf{0.17} & \textbf{0.15} & \textbf{0.20} & \textbf{0.14} & 0.15 \\
 & World & -- & -- & \textbf{0.15} & 0.15 & 0.15 & \textbf{0.46} \\
 & Sentiment & \textbf{0.15} & 0.15 & \textbf{0.15} & 0.15 & 0.15 & 0.16 \\
 & Alphanumeric & 0.16 & -- & 0.17 & -- & 0.15 & -- \\
\bottomrule
\end{tabular}%
}
\caption{Normalized cumulative regret at the final turn for Qwen3-32B, variance condition: No, by reward scale and nomenclature type. Scale labels: H\,=\,high reward magnitude; L\,=\,low; +\,=\,positive rewards; $-$\,=\,negative (e.g.\ H+ is the high positive scale).}
\label{tab:regret_Qwen3-32B_varNo}
\end{table}
\begin{table}[H]
\centering
\small
\resizebox{\columnwidth}{!}{%
\begin{tabular}{lr|rr|rr|rr}
\toprule
 &  & \multicolumn{2}{c}{Bandit} & \multicolumn{2}{c}{Farm} & \multicolumn{2}{c}{Clothing} \\
\cmidrule(lr){3-4} \cmidrule(lr){5-6} \cmidrule(lr){7-8}
Scale & Nomenclature & Helpful & Misleading & Helpful & Misleading & Helpful & Misleading \\
\midrule
H+ & Ordinal & \textbf{0.00} & \textbf{0.88} & \textbf{0.03} & \textbf{0.70} & \textbf{-0.01} & \textbf{0.89} \\
 & World & -- & -- & 0.11 & 0.26 & 0.04 & 0.81 \\
 & Sentiment & 0.12 & 0.54 & 0.16 & 0.28 & 0.09 & 0.37 \\
 & Alphanumeric & 0.21 & -- & 0.13 & -- & 0.25 & -- \\
\midrule
L+ & Ordinal & \textbf{0.00} & \textbf{0.72} & \textbf{0.04} & \textbf{0.51} & 0.05 & \textbf{0.77} \\
 & World & -- & -- & 0.10 & 0.21 & \textbf{0.04} & 0.58 \\
 & Sentiment & 0.13 & 0.31 & 0.16 & 0.34 & 0.11 & 0.29 \\
 & Alphanumeric & 0.21 & -- & 0.16 & -- & 0.17 & -- \\
\midrule
L- & Ordinal & 0.16 & \textbf{0.16} & \textbf{0.15} & \textbf{0.27} & 0.13 & 0.25 \\
 & World & -- & -- & 0.16 & 0.15 & 0.13 & \textbf{0.43} \\
 & Sentiment & \textbf{0.13} & 0.16 & 0.16 & 0.17 & \textbf{0.13} & 0.18 \\
 & Alphanumeric & 0.16 & -- & 0.17 & -- & 0.14 & -- \\
\midrule
H- & Ordinal & \textbf{0.14} & 0.13 & \textbf{0.15} & 0.17 & 0.15 & 0.19 \\
 & World & -- & -- & 0.16 & 0.15 & \textbf{0.12} & \textbf{0.40} \\
 & Sentiment & 0.15 & \textbf{0.14} & 0.18 & \textbf{0.18} & 0.16 & 0.15 \\
 & Alphanumeric & 0.16 & -- & 0.15 & -- & 0.15 & -- \\
\bottomrule
\end{tabular}%
}
\caption{Normalized cumulative regret at the final turn for Qwen3-32B, variance condition: Low, by reward scale and nomenclature type. Scale labels: H\,=\,high reward magnitude; L\,=\,low; +\,=\,positive rewards; $-$\,=\,negative (e.g.\ H+ is the high positive scale).}
\label{tab:regret_Qwen3-32B_varLow}
\end{table}
\begin{table}[H]
\centering
\small
\resizebox{\columnwidth}{!}{%
\begin{tabular}{lr|rr|rr|rr}
\toprule
 &  & \multicolumn{2}{c}{Bandit} & \multicolumn{2}{c}{Farm} & \multicolumn{2}{c}{Clothing} \\
\cmidrule(lr){3-4} \cmidrule(lr){5-6} \cmidrule(lr){7-8}
Scale & Nomenclature & Helpful & Misleading & Helpful & Misleading & Helpful & Misleading \\
\midrule
H+ & Ordinal & \textbf{0.01} & \textbf{0.87} & \textbf{0.05} & \textbf{0.75} & \textbf{-0.00} & \textbf{0.80} \\
 & World & -- & -- & 0.17 & 0.29 & 0.03 & 0.73 \\
 & Sentiment & 0.19 & 0.37 & 0.14 & 0.32 & 0.13 & 0.40 \\
 & Alphanumeric & 0.19 & -- & 0.21 & -- & 0.16 & -- \\
\midrule
L+ & Ordinal & \textbf{0.02} & \textbf{0.43} & \textbf{0.01} & \textbf{0.39} & 0.06 & 0.57 \\
 & World & -- & -- & 0.17 & 0.27 & \textbf{0.03} & \textbf{0.65} \\
 & Sentiment & 0.09 & 0.25 & 0.18 & 0.25 & 0.13 & 0.32 \\
 & Alphanumeric & 0.13 & -- & 0.13 & -- & 0.14 & -- \\
\midrule
L- & Ordinal & 0.13 & \textbf{0.26} & 0.20 & 0.18 & 0.12 & 0.31 \\
 & World & -- & -- & 0.16 & 0.18 & \textbf{0.10} & \textbf{0.49} \\
 & Sentiment & \textbf{0.12} & 0.22 & 0.18 & \textbf{0.22} & 0.18 & 0.15 \\
 & Alphanumeric & 0.16 & -- & \textbf{0.16} & -- & 0.17 & -- \\
\midrule
H- & Ordinal & \textbf{0.12} & 0.16 & \textbf{0.15} & \textbf{0.21} & 0.16 & 0.18 \\
 & World & -- & -- & 0.16 & 0.14 & \textbf{0.11} & \textbf{0.45} \\
 & Sentiment & 0.13 & \textbf{0.20} & 0.17 & 0.18 & 0.18 & 0.20 \\
 & Alphanumeric & 0.20 & -- & 0.17 & -- & 0.19 & -- \\
\bottomrule
\end{tabular}%
}
\caption{Normalized cumulative regret at the final turn for Qwen3-32B, variance condition: High, by reward scale and nomenclature type. Scale labels: H\,=\,high reward magnitude; L\,=\,low; +\,=\,positive rewards; $-$\,=\,negative (e.g.\ H+ is the high positive scale).}
\label{tab:regret_Qwen3-32B_varHigh}
\end{table}
\begin{table}[H]
\centering
\small
\resizebox{\columnwidth}{!}{%
\begin{tabular}{lr|rr|rr|rr}
\toprule
 &  & \multicolumn{2}{c}{Bandit} & \multicolumn{2}{c}{Farm} & \multicolumn{2}{c}{Clothing} \\
\cmidrule(lr){3-4} \cmidrule(lr){5-6} \cmidrule(lr){7-8}
Scale & Nomenclature & Helpful & Misleading & Helpful & Misleading & Helpful & Misleading \\
\midrule
H+ & Ordinal & \textbf{0.13} & \textbf{0.27} & \textbf{0.15} & \textbf{0.17} & 0.15 & 0.22 \\
 & World & -- & -- & \textbf{0.15} & 0.15 & \textbf{0.07} & \textbf{0.56} \\
 & Sentiment & 0.15 & 0.15 & \textbf{0.15} & 0.15 & 0.15 & 0.15 \\
 & Alphanumeric & 0.15 & -- & \textbf{0.15} & -- & 0.15 & -- \\
\midrule
L+ & Ordinal & \textbf{0.15} & \textbf{0.20} & \textbf{0.15} & \textbf{0.16} & 0.15 & 0.15 \\
 & World & -- & -- & \textbf{0.15} & 0.15 & \textbf{0.08} & \textbf{0.49} \\
 & Sentiment & \textbf{0.15} & 0.15 & \textbf{0.15} & 0.15 & 0.15 & 0.15 \\
 & Alphanumeric & \textbf{0.15} & -- & \textbf{0.15} & -- & 0.15 & -- \\
\midrule
L- & Ordinal & \textbf{0.15} & \textbf{0.19} & \textbf{0.15} & \textbf{0.18} & 0.15 & 0.16 \\
 & World & -- & -- & \textbf{0.15} & 0.15 & \textbf{0.10} & \textbf{0.36} \\
 & Sentiment & \textbf{0.15} & 0.16 & 0.15 & 0.15 & 0.15 & 0.16 \\
 & Alphanumeric & 0.15 & -- & 0.15 & -- & 0.16 & -- \\
\midrule
H- & Ordinal & 0.15 & \textbf{0.18} & 0.15 & \textbf{0.18} & 0.15 & 0.15 \\
 & World & -- & -- & \textbf{0.15} & 0.15 & \textbf{0.12} & \textbf{0.27} \\
 & Sentiment & \textbf{0.15} & 0.15 & \textbf{0.15} & 0.15 & 0.15 & 0.15 \\
 & Alphanumeric & \textbf{0.15} & -- & \textbf{0.15} & -- & 0.15 & -- \\
\bottomrule
\end{tabular}%
}
\caption{Normalized cumulative regret at the final turn for Gemini 3.1 Flash Lite, variance condition: No, by reward scale and nomenclature type. Scale labels: H\,=\,high reward magnitude; L\,=\,low; +\,=\,positive rewards; $-$\,=\,negative (e.g.\ H+ is the high positive scale).}
\label{tab:regret_Gemini_31_Flash_Lite_varNo}
\end{table}
\begin{table}[H]
\centering
\small
\resizebox{\columnwidth}{!}{%
\begin{tabular}{lr|rr|rr|rr}
\toprule
 &  & \multicolumn{2}{c}{Bandit} & \multicolumn{2}{c}{Farm} & \multicolumn{2}{c}{Clothing} \\
\cmidrule(lr){3-4} \cmidrule(lr){5-6} \cmidrule(lr){7-8}
Scale & Nomenclature & Helpful & Misleading & Helpful & Misleading & Helpful & Misleading \\
\midrule
H+ & Ordinal & -- & -- & -- & -- & 0.13 & 0.18 \\
 & World & -- & -- & -- & -- & \textbf{0.06} & \textbf{0.51} \\
 & Sentiment & \textbf{0.14} & \textbf{0.14} & \textbf{0.15} & \textbf{0.14} & 0.14 & 0.18 \\
 & Alphanumeric & -- & -- & -- & -- & 0.15 & -- \\
\midrule
L+ & Ordinal & -- & -- & -- & -- & 0.14 & 0.16 \\
 & World & -- & -- & -- & -- & \textbf{0.07} & \textbf{0.52} \\
 & Sentiment & \textbf{0.15} & \textbf{0.17} & \textbf{0.14} & \textbf{0.16} & 0.13 & 0.17 \\
 & Alphanumeric & -- & -- & -- & -- & 0.15 & -- \\
\midrule
L- & Ordinal & -- & -- & -- & -- & 0.15 & 0.16 \\
 & World & -- & -- & -- & -- & \textbf{0.11} & \textbf{0.24} \\
 & Sentiment & \textbf{0.14} & \textbf{0.14} & \textbf{0.17} & \textbf{0.16} & 0.14 & 0.16 \\
 & Alphanumeric & -- & -- & -- & -- & 0.16 & -- \\
\midrule
H- & Ordinal & -- & -- & -- & -- & 0.15 & 0.17 \\
 & World & -- & -- & -- & -- & \textbf{0.10} & \textbf{0.26} \\
 & Sentiment & \textbf{0.13} & \textbf{0.16} & \textbf{0.15} & \textbf{0.15} & 0.13 & 0.15 \\
 & Alphanumeric & -- & -- & -- & -- & 0.17 & -- \\
\bottomrule
\end{tabular}%
}
\caption{Normalized cumulative regret at the final turn for Gemini 3.1 Flash Lite, variance condition: Low, by reward scale and nomenclature type. Scale labels: H\,=\,high reward magnitude; L\,=\,low; +\,=\,positive rewards; $-$\,=\,negative (e.g.\ H+ is the high positive scale).}
\label{tab:regret_Gemini_31_Flash_Lite_varLow}
\end{table}
\begin{table}[H]
\centering
\small
\resizebox{\columnwidth}{!}{%
\begin{tabular}{lr|rr|rr|rr}
\toprule
 &  & \multicolumn{2}{c}{Bandit} & \multicolumn{2}{c}{Farm} & \multicolumn{2}{c}{Clothing} \\
\cmidrule(lr){3-4} \cmidrule(lr){5-6} \cmidrule(lr){7-8}
Scale & Nomenclature & Helpful & Misleading & Helpful & Misleading & Helpful & Misleading \\
\midrule
H+ & Ordinal & -- & -- & -- & -- & 0.12 & 0.23 \\
 & World & -- & -- & -- & -- & \textbf{0.06} & \textbf{0.53} \\
 & Sentiment & \textbf{0.17} & \textbf{0.11} & \textbf{0.11} & \textbf{0.18} & 0.20 & 0.22 \\
 & Alphanumeric & -- & -- & -- & -- & 0.16 & -- \\
\midrule
L+ & Ordinal & -- & -- & -- & -- & 0.13 & 0.12 \\
 & World & -- & -- & -- & -- & \textbf{0.08} & \textbf{0.50} \\
 & Sentiment & \textbf{0.12} & \textbf{0.18} & \textbf{0.20} & \textbf{0.15} & 0.16 & 0.15 \\
 & Alphanumeric & -- & -- & -- & -- & 0.18 & -- \\
\midrule
L- & Ordinal & -- & -- & -- & -- & 0.12 & 0.19 \\
 & World & -- & -- & -- & -- & 0.15 & \textbf{0.37} \\
 & Sentiment & \textbf{0.13} & \textbf{0.18} & \textbf{0.19} & \textbf{0.18} & 0.18 & 0.17 \\
 & Alphanumeric & -- & -- & -- & -- & \textbf{0.12} & -- \\
\midrule
H- & Ordinal & -- & -- & -- & -- & 0.18 & 0.16 \\
 & World & -- & -- & -- & -- & \textbf{0.13} & \textbf{0.23} \\
 & Sentiment & \textbf{0.18} & \textbf{0.16} & \textbf{0.13} & \textbf{0.23} & 0.21 & 0.15 \\
 & Alphanumeric & -- & -- & -- & -- & 0.14 & -- \\
\bottomrule
\end{tabular}%
}
\caption{Normalized cumulative regret at the final turn for Gemini 3.1 Flash Lite, variance condition: High, by reward scale and nomenclature type. Scale labels: H\,=\,high reward magnitude; L\,=\,low; +\,=\,positive rewards; $-$\,=\,negative (e.g.\ H+ is the high positive scale).}
\label{tab:regret_Gemini_31_Flash_Lite_varHigh}
\end{table}

\FloatBarrier
\subsection{Exploration Count}
\label{app:exp_count}
These tables show the average exploration count at turn 3 (the earliest step that full exploration is possible) and turn 10 for all nomenclatures, domains, models, and variance levels.
\begin{table}[H]
\centering
\small
\resizebox{\columnwidth}{!}{%
\begin{tabular}{lr|rrrr|rrrr|rrrr}
\toprule
 &  & \multicolumn{4}{c}{Bandit} & \multicolumn{4}{c}{Farm} & \multicolumn{4}{c}{Clothing} \\
\cmidrule(lr){3-6} \cmidrule(lr){7-10} \cmidrule(lr){11-14}
 &  & \multicolumn{2}{c}{Helpful} & \multicolumn{2}{c}{Misleading} & \multicolumn{2}{c}{Helpful} & \multicolumn{2}{c}{Misleading} & \multicolumn{2}{c}{Helpful} & \multicolumn{2}{c}{Misleading} \\
\cmidrule(lr){3-4} \cmidrule(lr){5-6} \cmidrule(lr){7-8} \cmidrule(lr){9-10} \cmidrule(lr){11-12} \cmidrule(lr){13-14}
Scale & Nomenclature & @3 & @10 & @3 & @10 & @3 & @10 & @3 & @10 & @3 & @10 & @3 & @10 \\
\midrule
H+ & Ordinal & 1.0 & 1.6 & 1.5 & 1.5 & 1.3 & \textbf{1.5} & 1.1 & 1.1 & 1.0 & 1.0 & 1.0 & 1.0 \\
 & World & -- & -- & -- & -- & 1.0 & 1.1 & 1.0 & 1.1 & 1.1 & \textbf{1.2} & 1.0 & \textbf{1.2} \\
 & Sentiment & 1.3 & 2.0 & 1.4 & \textbf{1.7} & 1.0 & 1.2 & 1.4 & \textbf{1.8} & 1.1 & 1.1 & 1.0 & 1.0 \\
 & Alphanumeric & 1.6 & \textbf{2.5} & -- & -- & 1.2 & 1.4 & -- & -- & 1.1 & \textbf{1.2} & -- & -- \\
\midrule
L+ & Ordinal & 1.8 & 2.1 & 1.6 & 2.2 & 1.2 & 1.6 & 1.2 & 2.0 & 1.0 & 1.0 & 1.1 & 1.2 \\
 & World & -- & -- & -- & -- & 1.2 & 1.4 & 1.4 & 2.4 & 1.1 & 1.2 & 1.4 & \textbf{1.7} \\
 & Sentiment & 1.9 & 2.5 & 1.8 & \textbf{2.5} & 1.2 & 1.9 & 2.2 & \textbf{2.6} & 1.4 & 1.8 & 1.3 & \textbf{1.7} \\
 & Alphanumeric & 2.1 & \textbf{2.8} & -- & -- & 1.7 & \textbf{2.7} & -- & -- & 1.6 & \textbf{2.0} & -- & -- \\
\midrule
L- & Ordinal & 2.3 & 2.7 & 2.8 & 2.8 & 2.0 & 2.2 & 2.4 & 2.5 & 1.7 & 1.9 & 1.9 & 2.1 \\
 & World & -- & -- & -- & -- & 2.6 & 2.7 & 2.7 & \textbf{3.0} & 1.7 & 2.3 & 2.1 & \textbf{2.8} \\
 & Sentiment & 2.7 & \textbf{3.0} & 2.9 & \textbf{3.0} & 2.5 & 2.9 & 2.2 & 2.9 & 2.1 & 2.3 & 2.3 & 2.7 \\
 & Alphanumeric & 3.0 & \textbf{3.0} & -- & -- & 2.4 & \textbf{3.0} & -- & -- & 2.6 & \textbf{2.9} & -- & -- \\
\midrule
H- & Ordinal & 2.4 & 2.9 & 2.4 & 2.8 & 2.3 & 2.5 & 2.2 & 2.4 & 2.0 & 2.0 & 2.3 & 2.5 \\
 & World & -- & -- & -- & -- & 2.2 & \textbf{2.9} & 2.3 & \textbf{2.8} & 2.2 & 2.6 & 2.1 & \textbf{2.7} \\
 & Sentiment & 2.6 & \textbf{3.0} & 2.6 & \textbf{3.0} & 2.5 & 2.8 & 2.3 & \textbf{2.8} & 2.3 & 2.3 & 2.1 & \textbf{2.7} \\
 & Alphanumeric & 2.8 & \textbf{3.0} & -- & -- & 2.4 & 2.8 & -- & -- & 2.1 & \textbf{2.8} & -- & -- \\
\midrule
H+ & UCB1 & 3.0 & 3.0 & -- & -- & 3.0 & 3.0 & -- & -- & 3.0 & 3.0 & -- & -- \\
\midrule
L+ & UCB1 & -- & -- & -- & -- & -- & -- & -- & -- & -- & -- & -- & -- \\
\midrule
L- & UCB1 & -- & -- & -- & -- & -- & -- & -- & -- & -- & -- & -- & -- \\
\midrule
H- & UCB1 & 3.0 & 3.0 & -- & -- & 3.0 & 3.0 & -- & -- & 3.0 & 3.0 & -- & -- \\
\bottomrule
\end{tabular}%
}
\caption{Exploration count (unique arms tried) at turns 3 and 10 for OLMo-3.1 32B, variance condition: No, by reward scale and nomenclature type. Scale labels: H\,=\,high reward magnitude; L\,=\,low; +\,=\,positive rewards; $-$\,=\,negative (e.g.\ H+ is the high positive scale).}
\label{tab:exploration_OLMo-31_32B_varNo}
\end{table}
\begin{table}[H]
\centering
\small
\resizebox{\columnwidth}{!}{%
\begin{tabular}{lr|rrrr|rrrr|rrrr}
\toprule
 &  & \multicolumn{4}{c}{Bandit} & \multicolumn{4}{c}{Farm} & \multicolumn{4}{c}{Clothing} \\
\cmidrule(lr){3-6} \cmidrule(lr){7-10} \cmidrule(lr){11-14}
 &  & \multicolumn{2}{c}{Helpful} & \multicolumn{2}{c}{Misleading} & \multicolumn{2}{c}{Helpful} & \multicolumn{2}{c}{Misleading} & \multicolumn{2}{c}{Helpful} & \multicolumn{2}{c}{Misleading} \\
\cmidrule(lr){3-4} \cmidrule(lr){5-6} \cmidrule(lr){7-8} \cmidrule(lr){9-10} \cmidrule(lr){11-12} \cmidrule(lr){13-14}
Scale & Nomenclature & @3 & @10 & @3 & @10 & @3 & @10 & @3 & @10 & @3 & @10 & @3 & @10 \\
\midrule
H+ & Ordinal & 1.6 & 1.9 & 1.4 & 1.4 & 1.3 & 1.3 & 1.0 & 1.3 & 1.0 & 1.0 & 1.1 & 1.2 \\
 & World & -- & -- & -- & -- & 1.1 & 1.4 & 1.3 & \textbf{2.0} & 1.1 & \textbf{1.5} & 1.1 & \textbf{1.4} \\
 & Sentiment & 1.9 & 1.9 & 1.8 & \textbf{2.1} & 1.4 & \textbf{1.8} & 1.5 & 1.7 & 1.0 & 1.1 & 1.0 & 1.0 \\
 & Alphanumeric & 1.8 & \textbf{2.2} & -- & -- & 1.3 & \textbf{1.8} & -- & -- & 1.0 & 1.0 & -- & -- \\
\midrule
L+ & Ordinal & 1.7 & 2.1 & 1.7 & 2.0 & 1.3 & 1.9 & 1.5 & 2.1 & 1.0 & 1.1 & 1.0 & 1.2 \\
 & World & -- & -- & -- & -- & 1.3 & 1.5 & 1.5 & 2.3 & 1.3 & \textbf{1.7} & 1.2 & \textbf{1.8} \\
 & Sentiment & 1.7 & 2.4 & 1.9 & \textbf{2.5} & 1.9 & \textbf{2.4} & 1.9 & \textbf{2.5} & 1.1 & 1.3 & 1.4 & \textbf{1.8} \\
 & Alphanumeric & 2.5 & \textbf{3.0} & -- & -- & 1.9 & \textbf{2.4} & -- & -- & 1.3 & 1.6 & -- & -- \\
\midrule
L- & Ordinal & 2.2 & 2.5 & 2.7 & 2.9 & 1.8 & 2.1 & 2.1 & 2.4 & 1.7 & 2.1 & 1.8 & 2.2 \\
 & World & -- & -- & -- & -- & 2.3 & 2.9 & 2.5 & 2.9 & 2.0 & 2.3 & 2.0 & \textbf{2.6} \\
 & Sentiment & 2.8 & \textbf{3.0} & 2.6 & \textbf{3.0} & 2.3 & 2.8 & 2.4 & \textbf{3.0} & 2.0 & 2.3 & 2.1 & \textbf{2.6} \\
 & Alphanumeric & 2.7 & \textbf{3.0} & -- & -- & 2.6 & \textbf{3.0} & -- & -- & 2.3 & \textbf{2.8} & -- & -- \\
\midrule
H- & Ordinal & 2.5 & 2.7 & 2.5 & 2.8 & 2.2 & 2.2 & 1.9 & 2.4 & 2.0 & 2.3 & 1.6 & 2.2 \\
 & World & -- & -- & -- & -- & 2.4 & \textbf{2.8} & 2.5 & \textbf{2.8} & 2.2 & 2.5 & 2.4 & \textbf{2.8} \\
 & Sentiment & 2.7 & 2.9 & 2.6 & \textbf{2.9} & 2.4 & \textbf{2.8} & 2.3 & 2.7 & 2.2 & 2.4 & 2.2 & 2.5 \\
 & Alphanumeric & 2.7 & \textbf{3.0} & -- & -- & 2.2 & 2.6 & -- & -- & 2.4 & \textbf{2.9} & -- & -- \\
\midrule
H+ & UCB1 & 3.0 & 3.0 & -- & -- & 3.0 & 3.0 & -- & -- & 3.0 & 3.0 & -- & -- \\
\midrule
L+ & UCB1 & -- & -- & -- & -- & -- & -- & -- & -- & -- & -- & -- & -- \\
\midrule
L- & UCB1 & -- & -- & -- & -- & -- & -- & -- & -- & -- & -- & -- & -- \\
\midrule
H- & UCB1 & 3.0 & 3.0 & -- & -- & 3.0 & 3.0 & -- & -- & 3.0 & 3.0 & -- & -- \\
\bottomrule
\end{tabular}%
}
\caption{Exploration count (unique arms tried) at turns 3 and 10 for OLMo-3.1 32B, variance condition: Low, by reward scale and nomenclature type. Scale labels: H\,=\,high reward magnitude; L\,=\,low; +\,=\,positive rewards; $-$\,=\,negative (e.g.\ H+ is the high positive scale).}
\label{tab:exploration_OLMo-31_32B_varLow}
\end{table}
\begin{table}[H]
\centering
\small
\resizebox{\columnwidth}{!}{%
\begin{tabular}{lr|rrrr|rrrr|rrrr}
\toprule
 &  & \multicolumn{4}{c}{Bandit} & \multicolumn{4}{c}{Farm} & \multicolumn{4}{c}{Clothing} \\
\cmidrule(lr){3-6} \cmidrule(lr){7-10} \cmidrule(lr){11-14}
 &  & \multicolumn{2}{c}{Helpful} & \multicolumn{2}{c}{Misleading} & \multicolumn{2}{c}{Helpful} & \multicolumn{2}{c}{Misleading} & \multicolumn{2}{c}{Helpful} & \multicolumn{2}{c}{Misleading} \\
\cmidrule(lr){3-4} \cmidrule(lr){5-6} \cmidrule(lr){7-8} \cmidrule(lr){9-10} \cmidrule(lr){11-12} \cmidrule(lr){13-14}
Scale & Nomenclature & @3 & @10 & @3 & @10 & @3 & @10 & @3 & @10 & @3 & @10 & @3 & @10 \\
\midrule
H+ & Ordinal & 1.6 & 1.7 & 1.3 & 1.4 & 1.2 & 1.2 & 1.0 & 1.2 & 1.0 & 1.1 & 1.1 & 1.3 \\
 & World & -- & -- & -- & -- & 1.0 & 1.3 & 1.0 & 1.2 & 1.0 & \textbf{1.3} & 1.2 & \textbf{1.5} \\
 & Sentiment & 1.3 & 1.8 & 1.4 & \textbf{1.5} & 1.4 & \textbf{1.9} & 1.3 & \textbf{1.6} & 1.0 & 1.0 & 1.2 & \textbf{1.5} \\
 & Alphanumeric & 1.8 & \textbf{2.6} & -- & -- & 1.2 & \textbf{1.9} & -- & -- & 1.0 & 1.1 & -- & -- \\
\midrule
L+ & Ordinal & 1.4 & 1.5 & 1.1 & 1.6 & 1.4 & 1.6 & 1.4 & 2.2 & 1.1 & 1.1 & 1.1 & 1.2 \\
 & World & -- & -- & -- & -- & 1.4 & 1.7 & 1.5 & 2.1 & 1.0 & 1.3 & 1.3 & \textbf{1.7} \\
 & Sentiment & 1.9 & 2.7 & 1.9 & \textbf{2.5} & 1.8 & 2.3 & 2.0 & \textbf{2.5} & 1.1 & 1.3 & 1.2 & 1.6 \\
 & Alphanumeric & 2.1 & \textbf{2.9} & -- & -- & 1.8 & \textbf{2.4} & -- & -- & 1.4 & \textbf{1.8} & -- & -- \\
\midrule
L- & Ordinal & 2.2 & 2.6 & 2.2 & 2.8 & 2.2 & 2.3 & 2.3 & 2.5 & 1.5 & 2.1 & 1.9 & 2.3 \\
 & World & -- & -- & -- & -- & 2.5 & \textbf{2.9} & 2.6 & \textbf{3.0} & 2.1 & 2.4 & 2.1 & \textbf{2.7} \\
 & Sentiment & 2.7 & \textbf{3.0} & 2.8 & \textbf{3.0} & 2.4 & \textbf{2.9} & 2.2 & 2.8 & 2.2 & 2.6 & 2.4 & 2.5 \\
 & Alphanumeric & 2.8 & \textbf{3.0} & -- & -- & 2.2 & \textbf{2.9} & -- & -- & 2.3 & \textbf{2.9} & -- & -- \\
\midrule
H- & Ordinal & 2.2 & 2.7 & 2.5 & 2.7 & 2.0 & 2.1 & 2.0 & 2.4 & 2.1 & 2.2 & 2.0 & 2.2 \\
 & World & -- & -- & -- & -- & 2.4 & \textbf{2.9} & 2.7 & \textbf{3.0} & 2.3 & \textbf{2.7} & 1.9 & \textbf{2.9} \\
 & Sentiment & 2.5 & \textbf{3.0} & 2.8 & \textbf{2.9} & 2.4 & 2.8 & 2.4 & 2.7 & 2.1 & 2.4 & 2.2 & 2.5 \\
 & Alphanumeric & 2.8 & \textbf{3.0} & -- & -- & 2.7 & \textbf{2.9} & -- & -- & 2.0 & 2.6 & -- & -- \\
\midrule
H+ & UCB1 & 3.0 & 3.0 & -- & -- & 3.0 & 3.0 & -- & -- & 3.0 & 3.0 & -- & -- \\
\midrule
L+ & UCB1 & -- & -- & -- & -- & -- & -- & -- & -- & -- & -- & -- & -- \\
\midrule
L- & UCB1 & -- & -- & -- & -- & -- & -- & -- & -- & -- & -- & -- & -- \\
\midrule
H- & UCB1 & 3.0 & 3.0 & -- & -- & 3.0 & 3.0 & -- & -- & 3.0 & 3.0 & -- & -- \\
\bottomrule
\end{tabular}%
}
\caption{Exploration count (unique arms tried) at turns 3 and 10 for OLMo-3.1 32B, variance condition: High, by reward scale and nomenclature type. Scale labels: H\,=\,high reward magnitude; L\,=\,low; +\,=\,positive rewards; $-$\,=\,negative (e.g.\ H+ is the high positive scale).}
\label{tab:exploration_OLMo-31_32B_varHigh}
\end{table}
\begin{table}[H]
\centering
\small
\resizebox{\columnwidth}{!}{%
\begin{tabular}{lr|rrrr|rrrr|rrrr}
\toprule
 &  & \multicolumn{4}{c}{Bandit} & \multicolumn{4}{c}{Farm} & \multicolumn{4}{c}{Clothing} \\
\cmidrule(lr){3-6} \cmidrule(lr){7-10} \cmidrule(lr){11-14}
 &  & \multicolumn{2}{c}{Helpful} & \multicolumn{2}{c}{Misleading} & \multicolumn{2}{c}{Helpful} & \multicolumn{2}{c}{Misleading} & \multicolumn{2}{c}{Helpful} & \multicolumn{2}{c}{Misleading} \\
\cmidrule(lr){3-4} \cmidrule(lr){5-6} \cmidrule(lr){7-8} \cmidrule(lr){9-10} \cmidrule(lr){11-12} \cmidrule(lr){13-14}
Scale & Nomenclature & @3 & @10 & @3 & @10 & @3 & @10 & @3 & @10 & @3 & @10 & @3 & @10 \\
\midrule
H+ & Ordinal & 1.0 & 1.0 & 1.1 & 1.7 & 1.0 & 1.6 & 1.2 & 2.1 & 1.2 & 1.2 & 1.1 & 1.3 \\
 & World & -- & -- & -- & -- & 1.9 & 2.5 & 2.0 & \textbf{2.9} & 1.3 & 1.7 & 1.2 & 1.4 \\
 & Sentiment & 1.9 & 2.2 & 2.4 & \textbf{2.8} & 2.1 & 2.4 & 2.6 & 2.7 & 2.0 & 2.2 & 2.3 & \textbf{2.4} \\
 & Alphanumeric & 2.8 & \textbf{3.0} & -- & -- & 2.7 & \textbf{3.0} & -- & -- & 2.6 & \textbf{3.0} & -- & -- \\
\midrule
L+ & Ordinal & 1.0 & 1.0 & 1.8 & \textbf{2.7} & 1.2 & 1.3 & 1.7 & 2.7 & 1.2 & 1.2 & 2.0 & \textbf{2.9} \\
 & World & -- & -- & -- & -- & 1.8 & 2.6 & 2.6 & \textbf{2.9} & 1.5 & 1.8 & 1.5 & 2.2 \\
 & Sentiment & 2.2 & \textbf{2.7} & 2.2 & 2.4 & 2.5 & 2.8 & 2.2 & \textbf{2.9} & 2.2 & 2.4 & 2.4 & 2.6 \\
 & Alphanumeric & 2.6 & 2.6 & -- & -- & 2.7 & \textbf{3.0} & -- & -- & 3.0 & \textbf{3.0} & -- & -- \\
\midrule
L- & Ordinal & 2.6 & \textbf{3.0} & 2.9 & \textbf{3.0} & 2.2 & \textbf{3.0} & 2.9 & \textbf{3.0} & 2.6 & 2.9 & 3.0 & \textbf{3.0} \\
 & World & -- & -- & -- & -- & 3.0 & \textbf{3.0} & 3.0 & \textbf{3.0} & 2.3 & 2.7 & 2.2 & 2.8 \\
 & Sentiment & 2.9 & \textbf{3.0} & 2.8 & \textbf{3.0} & 3.0 & \textbf{3.0} & 2.8 & \textbf{3.0} & 3.0 & \textbf{3.0} & 2.8 & \textbf{3.0} \\
 & Alphanumeric & 3.0 & \textbf{3.0} & -- & -- & 3.0 & \textbf{3.0} & -- & -- & 2.8 & \textbf{3.0} & -- & -- \\
\midrule
H- & Ordinal & 2.7 & \textbf{3.0} & 3.0 & \textbf{3.0} & 2.8 & \textbf{3.0} & 2.9 & \textbf{3.0} & 2.5 & 2.9 & 3.0 & \textbf{3.0} \\
 & World & -- & -- & -- & -- & 3.0 & \textbf{3.0} & 3.0 & \textbf{3.0} & 2.5 & \textbf{3.0} & 2.3 & 2.8 \\
 & Sentiment & 2.9 & \textbf{3.0} & 3.0 & \textbf{3.0} & 3.0 & \textbf{3.0} & 3.0 & \textbf{3.0} & 2.9 & \textbf{3.0} & 2.8 & \textbf{3.0} \\
 & Alphanumeric & 3.0 & \textbf{3.0} & -- & -- & 3.0 & \textbf{3.0} & -- & -- & 3.0 & \textbf{3.0} & -- & -- \\
\midrule
H+ & UCB1 & 3.0 & 3.0 & -- & -- & 3.0 & 3.0 & -- & -- & 3.0 & 3.0 & -- & -- \\
\midrule
L+ & UCB1 & -- & -- & -- & -- & -- & -- & -- & -- & -- & -- & -- & -- \\
\midrule
L- & UCB1 & -- & -- & -- & -- & -- & -- & -- & -- & -- & -- & -- & -- \\
\midrule
H- & UCB1 & 3.0 & 3.0 & -- & -- & 3.0 & 3.0 & -- & -- & 3.0 & 3.0 & -- & -- \\
\bottomrule
\end{tabular}%
}
\caption{Exploration count (unique arms tried) at turns 3 and 10 for Qwen3-32B, variance condition: No, by reward scale and nomenclature type. Scale labels: H\,=\,high reward magnitude; L\,=\,low; +\,=\,positive rewards; $-$\,=\,negative (e.g.\ H+ is the high positive scale).}
\label{tab:exploration_Qwen3-32B_varNo}
\end{table}
\begin{table}[H]
\centering
\small
\resizebox{\columnwidth}{!}{%
\begin{tabular}{lr|rrrr|rrrr|rrrr}
\toprule
 &  & \multicolumn{4}{c}{Bandit} & \multicolumn{4}{c}{Farm} & \multicolumn{4}{c}{Clothing} \\
\cmidrule(lr){3-6} \cmidrule(lr){7-10} \cmidrule(lr){11-14}
 &  & \multicolumn{2}{c}{Helpful} & \multicolumn{2}{c}{Misleading} & \multicolumn{2}{c}{Helpful} & \multicolumn{2}{c}{Misleading} & \multicolumn{2}{c}{Helpful} & \multicolumn{2}{c}{Misleading} \\
\cmidrule(lr){3-4} \cmidrule(lr){5-6} \cmidrule(lr){7-8} \cmidrule(lr){9-10} \cmidrule(lr){11-12} \cmidrule(lr){13-14}
Scale & Nomenclature & @3 & @10 & @3 & @10 & @3 & @10 & @3 & @10 & @3 & @10 & @3 & @10 \\
\midrule
H+ & Ordinal & 1.2 & 1.3 & 1.1 & 1.5 & 1.3 & 1.5 & 1.4 & 1.8 & 1.2 & 1.3 & 1.2 & 1.4 \\
 & World & -- & -- & -- & -- & 1.9 & 2.4 & 2.2 & \textbf{2.8} & 1.3 & 1.5 & 1.2 & 1.5 \\
 & Sentiment & 1.5 & 2.1 & 1.7 & \textbf{2.1} & 2.0 & 2.4 & 2.6 & 2.7 & 1.7 & 1.9 & 2.1 & \textbf{2.4} \\
 & Alphanumeric & 2.5 & \textbf{2.7} & -- & -- & 3.0 & \textbf{3.0} & -- & -- & 2.4 & \textbf{2.7} & -- & -- \\
\midrule
L+ & Ordinal & 1.0 & 1.1 & 1.2 & 2.0 & 1.3 & 1.8 & 1.4 & 2.5 & 1.6 & 1.7 & 1.5 & 1.6 \\
 & World & -- & -- & -- & -- & 1.9 & 2.5 & 2.3 & \textbf{3.0} & 1.7 & 1.8 & 1.7 & 2.2 \\
 & Sentiment & 2.0 & 2.5 & 2.3 & \textbf{2.7} & 1.9 & 2.5 & 2.3 & 2.7 & 2.0 & 2.1 & 2.4 & \textbf{2.6} \\
 & Alphanumeric & 2.3 & \textbf{2.9} & -- & -- & 2.7 & \textbf{3.0} & -- & -- & 2.7 & \textbf{3.0} & -- & -- \\
\midrule
L- & Ordinal & 2.6 & 2.9 & 3.0 & \textbf{3.0} & 2.8 & \textbf{3.0} & 2.7 & \textbf{3.0} & 2.8 & 2.9 & 2.8 & \textbf{3.0} \\
 & World & -- & -- & -- & -- & 2.9 & \textbf{3.0} & 3.0 & \textbf{3.0} & 2.0 & 2.8 & 2.5 & 2.9 \\
 & Sentiment & 2.7 & 2.9 & 2.9 & \textbf{3.0} & 3.0 & \textbf{3.0} & 2.8 & \textbf{3.0} & 2.8 & 2.9 & 2.8 & \textbf{3.0} \\
 & Alphanumeric & 3.0 & \textbf{3.0} & -- & -- & 3.0 & \textbf{3.0} & -- & -- & 2.9 & \textbf{3.0} & -- & -- \\
\midrule
H- & Ordinal & 2.6 & 2.9 & 3.0 & \textbf{3.0} & 2.8 & 2.9 & 3.0 & \textbf{3.0} & 2.7 & 2.9 & 2.9 & 2.9 \\
 & World & -- & -- & -- & -- & 3.0 & \textbf{3.0} & 2.9 & \textbf{3.0} & 2.5 & 2.9 & 2.7 & 2.8 \\
 & Sentiment & 2.8 & \textbf{3.0} & 2.9 & \textbf{3.0} & 2.9 & \textbf{3.0} & 2.9 & \textbf{3.0} & 2.8 & \textbf{3.0} & 3.0 & \textbf{3.0} \\
 & Alphanumeric & 3.0 & \textbf{3.0} & -- & -- & 3.0 & \textbf{3.0} & -- & -- & 3.0 & \textbf{3.0} & -- & -- \\
\midrule
H+ & UCB1 & 3.0 & 3.0 & -- & -- & 3.0 & 3.0 & -- & -- & 3.0 & 3.0 & -- & -- \\
\midrule
L+ & UCB1 & -- & -- & -- & -- & -- & -- & -- & -- & -- & -- & -- & -- \\
\midrule
L- & UCB1 & -- & -- & -- & -- & -- & -- & -- & -- & -- & -- & -- & -- \\
\midrule
H- & UCB1 & 3.0 & 3.0 & -- & -- & 3.0 & 3.0 & -- & -- & 3.0 & 3.0 & -- & -- \\
\bottomrule
\end{tabular}%
}
\caption{Exploration count (unique arms tried) at turns 3 and 10 for Qwen3-32B, variance condition: Low, by reward scale and nomenclature type. Scale labels: H\,=\,high reward magnitude; L\,=\,low; +\,=\,positive rewards; $-$\,=\,negative (e.g.\ H+ is the high positive scale).}
\label{tab:exploration_Qwen3-32B_varLow}
\end{table}
\begin{table}[H]
\centering
\small
\resizebox{\columnwidth}{!}{%
\begin{tabular}{lr|rrrr|rrrr|rrrr}
\toprule
 &  & \multicolumn{4}{c}{Bandit} & \multicolumn{4}{c}{Farm} & \multicolumn{4}{c}{Clothing} \\
\cmidrule(lr){3-6} \cmidrule(lr){7-10} \cmidrule(lr){11-14}
 &  & \multicolumn{2}{c}{Helpful} & \multicolumn{2}{c}{Misleading} & \multicolumn{2}{c}{Helpful} & \multicolumn{2}{c}{Misleading} & \multicolumn{2}{c}{Helpful} & \multicolumn{2}{c}{Misleading} \\
\cmidrule(lr){3-4} \cmidrule(lr){5-6} \cmidrule(lr){7-8} \cmidrule(lr){9-10} \cmidrule(lr){11-12} \cmidrule(lr){13-14}
Scale & Nomenclature & @3 & @10 & @3 & @10 & @3 & @10 & @3 & @10 & @3 & @10 & @3 & @10 \\
\midrule
H+ & Ordinal & 1.1 & 1.5 & 1.4 & 1.4 & 1.2 & 1.8 & 1.2 & 1.8 & 1.1 & 1.2 & 1.3 & 1.4 \\
 & World & -- & -- & -- & -- & 2.0 & 2.7 & 2.0 & \textbf{3.0} & 1.1 & 1.5 & 1.6 & 1.7 \\
 & Sentiment & 2.2 & 2.4 & 2.0 & \textbf{2.4} & 2.2 & 2.7 & 2.6 & 2.7 & 2.2 & 2.2 & 2.3 & \textbf{2.4} \\
 & Alphanumeric & 2.2 & \textbf{2.7} & -- & -- & 2.7 & \textbf{2.9} & -- & -- & 2.6 & \textbf{2.8} & -- & -- \\
\midrule
L+ & Ordinal & 1.0 & 1.5 & 1.8 & 2.6 & 1.2 & 1.6 & 1.9 & 2.8 & 1.1 & 1.4 & 1.7 & 2.1 \\
 & World & -- & -- & -- & -- & 2.1 & 2.7 & 2.2 & \textbf{2.9} & 1.4 & 1.5 & 1.5 & 2.1 \\
 & Sentiment & 2.0 & 2.4 & 2.6 & \textbf{2.8} & 2.1 & 2.7 & 2.3 & 2.8 & 2.1 & 2.3 & 2.5 & \textbf{2.6} \\
 & Alphanumeric & 2.8 & \textbf{3.0} & -- & -- & 2.7 & \textbf{3.0} & -- & -- & 2.9 & \textbf{3.0} & -- & -- \\
\midrule
L- & Ordinal & 2.5 & 2.8 & 2.9 & 2.9 & 2.9 & \textbf{3.0} & 2.8 & \textbf{3.0} & 2.6 & 2.9 & 2.8 & 2.9 \\
 & World & -- & -- & -- & -- & 3.0 & \textbf{3.0} & 3.0 & \textbf{3.0} & 2.3 & 2.8 & 2.4 & 2.8 \\
 & Sentiment & 3.0 & \textbf{3.0} & 2.9 & \textbf{3.0} & 2.7 & \textbf{3.0} & 3.0 & \textbf{3.0} & 2.9 & \textbf{3.0} & 2.8 & \textbf{3.0} \\
 & Alphanumeric & 3.0 & \textbf{3.0} & -- & -- & 3.0 & \textbf{3.0} & -- & -- & 3.0 & \textbf{3.0} & -- & -- \\
\midrule
H- & Ordinal & 2.6 & 2.8 & 3.0 & \textbf{3.0} & 2.9 & \textbf{3.0} & 2.9 & \textbf{3.0} & 2.6 & 2.8 & 2.9 & \textbf{3.0} \\
 & World & -- & -- & -- & -- & 3.0 & \textbf{3.0} & 3.0 & \textbf{3.0} & 2.2 & 2.8 & 2.6 & 2.9 \\
 & Sentiment & 3.0 & \textbf{3.0} & 3.0 & \textbf{3.0} & 3.0 & \textbf{3.0} & 3.0 & \textbf{3.0} & 3.0 & \textbf{3.0} & 2.9 & \textbf{3.0} \\
 & Alphanumeric & 3.0 & \textbf{3.0} & -- & -- & 3.0 & \textbf{3.0} & -- & -- & 3.0 & \textbf{3.0} & -- & -- \\
\midrule
H+ & UCB1 & 3.0 & 3.0 & -- & -- & 3.0 & 3.0 & -- & -- & 3.0 & 3.0 & -- & -- \\
\midrule
L+ & UCB1 & -- & -- & -- & -- & -- & -- & -- & -- & -- & -- & -- & -- \\
\midrule
L- & UCB1 & -- & -- & -- & -- & -- & -- & -- & -- & -- & -- & -- & -- \\
\midrule
H- & UCB1 & 3.0 & 3.0 & -- & -- & 3.0 & 3.0 & -- & -- & 3.0 & 3.0 & -- & -- \\
\bottomrule
\end{tabular}%
}
\caption{Exploration count (unique arms tried) at turns 3 and 10 for Qwen3-32B, variance condition: High, by reward scale and nomenclature type. Scale labels: H\,=\,high reward magnitude; L\,=\,low; +\,=\,positive rewards; $-$\,=\,negative (e.g.\ H+ is the high positive scale).}
\label{tab:exploration_Qwen3-32B_varHigh}
\end{table}
\begin{table}[H]
\centering
\small
\resizebox{\columnwidth}{!}{%
\begin{tabular}{lr|rrrr|rrrr|rrrr}
\toprule
 &  & \multicolumn{4}{c}{Bandit} & \multicolumn{4}{c}{Farm} & \multicolumn{4}{c}{Clothing} \\
\cmidrule(lr){3-6} \cmidrule(lr){7-10} \cmidrule(lr){11-14}
 &  & \multicolumn{2}{c}{Helpful} & \multicolumn{2}{c}{Misleading} & \multicolumn{2}{c}{Helpful} & \multicolumn{2}{c}{Misleading} & \multicolumn{2}{c}{Helpful} & \multicolumn{2}{c}{Misleading} \\
\cmidrule(lr){3-4} \cmidrule(lr){5-6} \cmidrule(lr){7-8} \cmidrule(lr){9-10} \cmidrule(lr){11-12} \cmidrule(lr){13-14}
Scale & Nomenclature & @3 & @10 & @3 & @10 & @3 & @10 & @3 & @10 & @3 & @10 & @3 & @10 \\
\midrule
H+ & Ordinal & 2.8 & 2.8 & 2.4 & 2.9 & 2.7 & \textbf{3.0} & 3.0 & \textbf{3.0} & 2.6 & \textbf{3.0} & 2.6 & 2.9 \\
 & World & -- & -- & -- & -- & 3.0 & \textbf{3.0} & 3.0 & \textbf{3.0} & 1.9 & 2.1 & 1.9 & 2.2 \\
 & Sentiment & 3.0 & \textbf{3.0} & 3.0 & \textbf{3.0} & 3.0 & \textbf{3.0} & 3.0 & \textbf{3.0} & 3.0 & \textbf{3.0} & 3.0 & \textbf{3.0} \\
 & Alphanumeric & 3.0 & \textbf{3.0} & -- & -- & 3.0 & \textbf{3.0} & -- & -- & 3.0 & \textbf{3.0} & -- & -- \\
\midrule
L+ & Ordinal & 3.0 & \textbf{3.0} & 3.0 & \textbf{3.0} & 3.0 & \textbf{3.0} & 3.0 & \textbf{3.0} & 2.7 & \textbf{3.0} & 3.0 & \textbf{3.0} \\
 & World & -- & -- & -- & -- & 3.0 & \textbf{3.0} & 3.0 & \textbf{3.0} & 2.1 & 2.3 & 2.0 & 2.4 \\
 & Sentiment & 3.0 & \textbf{3.0} & 3.0 & \textbf{3.0} & 3.0 & \textbf{3.0} & 2.9 & \textbf{3.0} & 3.0 & \textbf{3.0} & 3.0 & \textbf{3.0} \\
 & Alphanumeric & 3.0 & \textbf{3.0} & -- & -- & 3.0 & \textbf{3.0} & -- & -- & 3.0 & \textbf{3.0} & -- & -- \\
\midrule
L- & Ordinal & 2.9 & \textbf{3.0} & 3.0 & \textbf{3.0} & 3.0 & \textbf{3.0} & 3.0 & \textbf{3.0} & 2.7 & \textbf{3.0} & 3.0 & \textbf{3.0} \\
 & World & -- & -- & -- & -- & 3.0 & \textbf{3.0} & 3.0 & \textbf{3.0} & 2.1 & 2.5 & 2.4 & 2.9 \\
 & Sentiment & 3.0 & \textbf{3.0} & 3.0 & \textbf{3.0} & 3.0 & \textbf{3.0} & 3.0 & \textbf{3.0} & 3.0 & \textbf{3.0} & 3.0 & \textbf{3.0} \\
 & Alphanumeric & 3.0 & \textbf{3.0} & -- & -- & 3.0 & \textbf{3.0} & -- & -- & 3.0 & \textbf{3.0} & -- & -- \\
\midrule
H- & Ordinal & 3.0 & \textbf{3.0} & 3.0 & \textbf{3.0} & 3.0 & \textbf{3.0} & 3.0 & \textbf{3.0} & 3.0 & \textbf{3.0} & 3.0 & \textbf{3.0} \\
 & World & -- & -- & -- & -- & 3.0 & \textbf{3.0} & 3.0 & \textbf{3.0} & 2.3 & 2.7 & 2.6 & \textbf{3.0} \\
 & Sentiment & 3.0 & \textbf{3.0} & 3.0 & \textbf{3.0} & 3.0 & \textbf{3.0} & 3.0 & \textbf{3.0} & 3.0 & \textbf{3.0} & 3.0 & \textbf{3.0} \\
 & Alphanumeric & 3.0 & \textbf{3.0} & -- & -- & 3.0 & \textbf{3.0} & -- & -- & 3.0 & \textbf{3.0} & -- & -- \\
\midrule
H+ & UCB1 & 3.0 & 3.0 & -- & -- & 3.0 & 3.0 & -- & -- & 3.0 & 3.0 & -- & -- \\
\midrule
L+ & UCB1 & -- & -- & -- & -- & -- & -- & -- & -- & -- & -- & -- & -- \\
\midrule
L- & UCB1 & -- & -- & -- & -- & -- & -- & -- & -- & -- & -- & -- & -- \\
\midrule
H- & UCB1 & 3.0 & 3.0 & -- & -- & 3.0 & 3.0 & -- & -- & 3.0 & 3.0 & -- & -- \\
\bottomrule
\end{tabular}%
}
\caption{Exploration count (unique arms tried) at turns 3 and 10 for Gemini 3.1 Flash Lite, variance condition: No, by reward scale and nomenclature type. Scale labels: H\,=\,high reward magnitude; L\,=\,low; +\,=\,positive rewards; $-$\,=\,negative (e.g.\ H+ is the high positive scale).}
\label{tab:exploration_Gemini_31_Flash_Lite_varNo}
\end{table}
\begin{table}[H]
\centering
\small
\resizebox{\columnwidth}{!}{%
\begin{tabular}{lr|rrrr|rrrr|rrrr}
\toprule
 &  & \multicolumn{4}{c}{Bandit} & \multicolumn{4}{c}{Farm} & \multicolumn{4}{c}{Clothing} \\
\cmidrule(lr){3-6} \cmidrule(lr){7-10} \cmidrule(lr){11-14}
 &  & \multicolumn{2}{c}{Helpful} & \multicolumn{2}{c}{Misleading} & \multicolumn{2}{c}{Helpful} & \multicolumn{2}{c}{Misleading} & \multicolumn{2}{c}{Helpful} & \multicolumn{2}{c}{Misleading} \\
\cmidrule(lr){3-4} \cmidrule(lr){5-6} \cmidrule(lr){7-8} \cmidrule(lr){9-10} \cmidrule(lr){11-12} \cmidrule(lr){13-14}
Scale & Nomenclature & @3 & @10 & @3 & @10 & @3 & @10 & @3 & @10 & @3 & @10 & @3 & @10 \\
\midrule
H+ & Ordinal & -- & -- & -- & -- & -- & -- & -- & -- & 2.8 & \textbf{3.0} & 2.7 & \textbf{3.0} \\
 & World & -- & -- & -- & -- & -- & -- & -- & -- & 1.9 & 2.0 & 2.0 & 2.2 \\
 & Sentiment & 3.0 & \textbf{3.0} & 2.9 & \textbf{3.0} & 3.0 & \textbf{3.0} & 3.0 & \textbf{3.0} & 2.9 & \textbf{3.0} & 2.9 & \textbf{3.0} \\
 & Alphanumeric & -- & -- & -- & -- & -- & -- & -- & -- & 3.0 & \textbf{3.0} & -- & -- \\
\midrule
L+ & Ordinal & -- & -- & -- & -- & -- & -- & -- & -- & 2.8 & \textbf{3.0} & 3.0 & \textbf{3.0} \\
 & World & -- & -- & -- & -- & -- & -- & -- & -- & 1.8 & 2.2 & 1.9 & 2.4 \\
 & Sentiment & 3.0 & \textbf{3.0} & 3.0 & \textbf{3.0} & 3.0 & \textbf{3.0} & 3.0 & \textbf{3.0} & 2.8 & \textbf{3.0} & 3.0 & \textbf{3.0} \\
 & Alphanumeric & -- & -- & -- & -- & -- & -- & -- & -- & 3.0 & \textbf{3.0} & -- & -- \\
\midrule
L- & Ordinal & -- & -- & -- & -- & -- & -- & -- & -- & 2.9 & \textbf{3.0} & 3.0 & \textbf{3.0} \\
 & World & -- & -- & -- & -- & -- & -- & -- & -- & 2.2 & 2.6 & 3.0 & \textbf{3.0} \\
 & Sentiment & 3.0 & \textbf{3.0} & 3.0 & \textbf{3.0} & 3.0 & \textbf{3.0} & 3.0 & \textbf{3.0} & 3.0 & \textbf{3.0} & 3.0 & \textbf{3.0} \\
 & Alphanumeric & -- & -- & -- & -- & -- & -- & -- & -- & 3.0 & \textbf{3.0} & -- & -- \\
\midrule
H- & Ordinal & -- & -- & -- & -- & -- & -- & -- & -- & 2.9 & \textbf{3.0} & 3.0 & \textbf{3.0} \\
 & World & -- & -- & -- & -- & -- & -- & -- & -- & 2.3 & 2.5 & 2.4 & 2.8 \\
 & Sentiment & 3.0 & \textbf{3.0} & 3.0 & \textbf{3.0} & 3.0 & \textbf{3.0} & 3.0 & \textbf{3.0} & 3.0 & \textbf{3.0} & 3.0 & \textbf{3.0} \\
 & Alphanumeric & -- & -- & -- & -- & -- & -- & -- & -- & 3.0 & \textbf{3.0} & -- & -- \\
\midrule
H+ & UCB1 & 3.0 & 3.0 & -- & -- & 3.0 & 3.0 & -- & -- & 3.0 & 3.0 & -- & -- \\
\midrule
L+ & UCB1 & -- & -- & -- & -- & -- & -- & -- & -- & -- & -- & -- & -- \\
\midrule
L- & UCB1 & -- & -- & -- & -- & -- & -- & -- & -- & -- & -- & -- & -- \\
\midrule
H- & UCB1 & 3.0 & 3.0 & -- & -- & 3.0 & 3.0 & -- & -- & 3.0 & 3.0 & -- & -- \\
\bottomrule
\end{tabular}%
}
\caption{Exploration count (unique arms tried) at turns 3 and 10 for Gemini 3.1 Flash Lite, variance condition: Low, by reward scale and nomenclature type. Scale labels: H\,=\,high reward magnitude; L\,=\,low; +\,=\,positive rewards; $-$\,=\,negative (e.g.\ H+ is the high positive scale).}
\label{tab:exploration_Gemini_31_Flash_Lite_varLow}
\end{table}
\begin{table}[H]
\centering
\small
\resizebox{\columnwidth}{!}{%
\begin{tabular}{lr|rrrr|rrrr|rrrr}
\toprule
 &  & \multicolumn{4}{c}{Bandit} & \multicolumn{4}{c}{Farm} & \multicolumn{4}{c}{Clothing} \\
\cmidrule(lr){3-6} \cmidrule(lr){7-10} \cmidrule(lr){11-14}
 &  & \multicolumn{2}{c}{Helpful} & \multicolumn{2}{c}{Misleading} & \multicolumn{2}{c}{Helpful} & \multicolumn{2}{c}{Misleading} & \multicolumn{2}{c}{Helpful} & \multicolumn{2}{c}{Misleading} \\
\cmidrule(lr){3-4} \cmidrule(lr){5-6} \cmidrule(lr){7-8} \cmidrule(lr){9-10} \cmidrule(lr){11-12} \cmidrule(lr){13-14}
Scale & Nomenclature & @3 & @10 & @3 & @10 & @3 & @10 & @3 & @10 & @3 & @10 & @3 & @10 \\
\midrule
H+ & Ordinal & -- & -- & -- & -- & -- & -- & -- & -- & 2.4 & 2.9 & 2.5 & 2.9 \\
 & World & -- & -- & -- & -- & -- & -- & -- & -- & 2.0 & 2.1 & 1.8 & 2.4 \\
 & Sentiment & 2.9 & \textbf{3.0} & 2.9 & \textbf{3.0} & 3.0 & \textbf{3.0} & 2.9 & \textbf{3.0} & 2.9 & \textbf{3.0} & 3.0 & \textbf{3.0} \\
 & Alphanumeric & -- & -- & -- & -- & -- & -- & -- & -- & 3.0 & \textbf{3.0} & -- & -- \\
\midrule
L+ & Ordinal & -- & -- & -- & -- & -- & -- & -- & -- & 3.0 & \textbf{3.0} & 2.9 & \textbf{3.0} \\
 & World & -- & -- & -- & -- & -- & -- & -- & -- & 2.1 & 2.2 & 1.9 & 2.4 \\
 & Sentiment & 3.0 & \textbf{3.0} & 3.0 & \textbf{3.0} & 3.0 & \textbf{3.0} & 3.0 & \textbf{3.0} & 3.0 & \textbf{3.0} & 3.0 & \textbf{3.0} \\
 & Alphanumeric & -- & -- & -- & -- & -- & -- & -- & -- & 3.0 & \textbf{3.0} & -- & -- \\
\midrule
L- & Ordinal & -- & -- & -- & -- & -- & -- & -- & -- & 3.0 & \textbf{3.0} & 3.0 & \textbf{3.0} \\
 & World & -- & -- & -- & -- & -- & -- & -- & -- & 2.1 & 2.9 & 2.5 & 2.9 \\
 & Sentiment & 3.0 & \textbf{3.0} & 3.0 & \textbf{3.0} & 3.0 & \textbf{3.0} & 3.0 & \textbf{3.0} & 3.0 & \textbf{3.0} & 3.0 & \textbf{3.0} \\
 & Alphanumeric & -- & -- & -- & -- & -- & -- & -- & -- & 3.0 & \textbf{3.0} & -- & -- \\
\midrule
H- & Ordinal & -- & -- & -- & -- & -- & -- & -- & -- & 3.0 & \textbf{3.0} & 3.0 & \textbf{3.0} \\
 & World & -- & -- & -- & -- & -- & -- & -- & -- & 2.2 & 2.8 & 2.4 & 2.9 \\
 & Sentiment & 3.0 & \textbf{3.0} & 3.0 & \textbf{3.0} & 3.0 & \textbf{3.0} & 3.0 & \textbf{3.0} & 3.0 & \textbf{3.0} & 3.0 & \textbf{3.0} \\
 & Alphanumeric & -- & -- & -- & -- & -- & -- & -- & -- & 3.0 & \textbf{3.0} & -- & -- \\
\midrule
H+ & UCB1 & 3.0 & 3.0 & -- & -- & 3.0 & 3.0 & -- & -- & 3.0 & 3.0 & -- & -- \\
\midrule
L+ & UCB1 & -- & -- & -- & -- & -- & -- & -- & -- & -- & -- & -- & -- \\
\midrule
L- & UCB1 & -- & -- & -- & -- & -- & -- & -- & -- & -- & -- & -- & -- \\
\midrule
H- & UCB1 & 3.0 & 3.0 & -- & -- & 3.0 & 3.0 & -- & -- & 3.0 & 3.0 & -- & -- \\
\bottomrule
\end{tabular}%
}
\caption{Exploration count (unique arms tried) at turns 3 and 10 for Gemini 3.1 Flash Lite, variance condition: High, by reward scale and nomenclature type. Scale labels: H\,=\,high reward magnitude; L\,=\,low; +\,=\,positive rewards; $-$\,=\,negative (e.g.\ H+ is the high positive scale).}
\label{tab:exploration_Gemini_31_Flash_Lite_varHigh}
\end{table}

\begin{figure*}[ht]
  \begin{center}
    \centerline{\includegraphics[width=\linewidth]{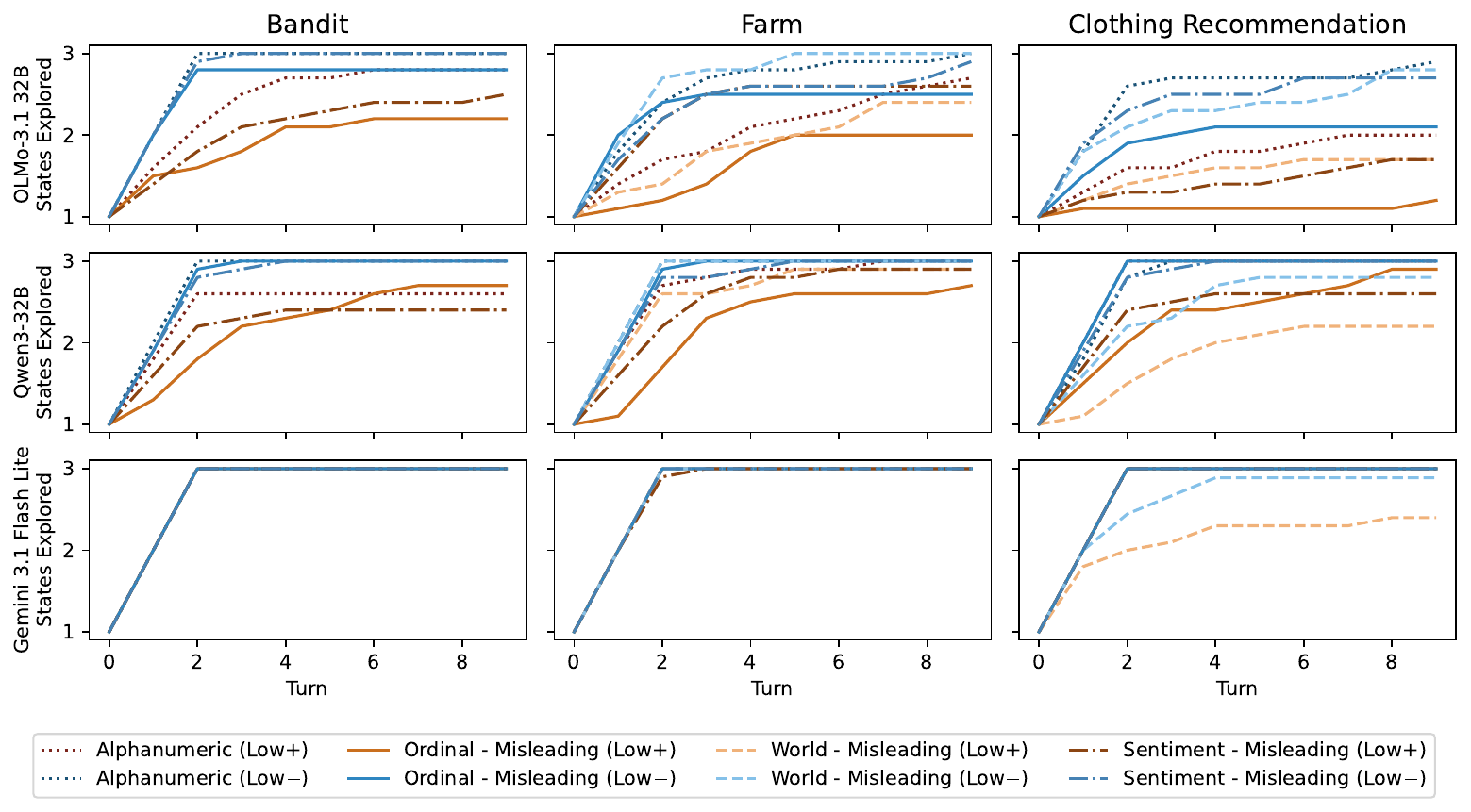}}
  \end{center}
  \caption{
      \textbf{Impact of Reward Scale on Exploration Count} We observe more exploration when reward is negative (blue) than positive (brown). These results are with no variance and \textbf{low scales}.
    \label{fig:scale_explore_low}
    }
 \end{figure*}

\FloatBarrier
% \subsection{Burst of Exploration}
% \input{figures/tables/full_exploration_table_Olmo-3.1-32B-Instruct_varNo} 
% \input{figures/tables/full_exploration_table_Olmo-3.1-32B-Instruct_varLow} 
% \input{figures/tables/full_exploration_table_Olmo-3.1-32B-Instruct_varHigh} 
% \input{figures/tables/full_exploration_table_Qwen3-32B_varNo} 
% \input{figures/tables/full_exploration_table_Qwen3-32B_varLow} 
% \input{figures/tables/full_exploration_table_Qwen3-32B_varHigh} 
% \input{figures/tables/full_exploration_table_gemini-3.1-flash-lite-preview_varNo} 
% \input{figures/tables/full_exploration_table_gemini-3.1-flash-lite-preview_varLow} 
% \input{figures/tables/full_exploration_table_gemini-3.1-flash-lite-preview_varHigh} 

\subsection{Qwen Scaling Results}
\label{app:qwen_scaling}
\begin{table}[ht]
\centering
\small
\resizebox{\linewidth}{!}{%
\begin{tabular}{ll|rrrr|rrrr|rrrr}
\toprule
  &   & \multicolumn{4}{c|}{Bandit} & \multicolumn{4}{c|}{Farm} & \multicolumn{4}{c}{Clothing} \\
\cmidrule(lr){3-6} \cmidrule(lr){7-10} \cmidrule(lr){11-14}
  &   & \multicolumn{2}{c}{Regret} & \multicolumn{2}{c|}{Exploration} & \multicolumn{2}{c}{Regret} & \multicolumn{2}{c|}{Exploration} & \multicolumn{2}{c}{Regret} & \multicolumn{2}{c}{Exploration} \\
\cmidrule(lr){3-4} \cmidrule(lr){5-6} \cmidrule(lr){7-8} \cmidrule(lr){9-10} \cmidrule(lr){11-12} \cmidrule(lr){13-14}
Model & Nomenclature & H & M & H & M & H & M & H & M & H & M & H & M \\
\midrule
Qwen3-8B & Ordinal & 0.01 & 0.95 & 1.10 & 1.10 & 0.00 & 1.00 & 1.00 & 1.00 & -- & -- & -- & -- \\
 & World & -- & -- & -- & -- & 0.01 & 0.78 & 1.10 & 1.50 & -- & -- & -- & -- \\
 & Sentiment & 0.33 & 0.60 & 1.10 & 1.00 & 0.20 & 0.70 & 1.00 & 1.00 & -- & -- & -- & -- \\
 & Alphanumeric & 0.40 & -- & 1.60 & -- & 0.34 & -- & 1.60 & -- & -- & -- & -- & -- \\
\cmidrule(lr){2-14}
Qwen3-14B & Ordinal & 0.07 & 0.49 & 1.70 & 2.30 & 0.04 & 0.71 & 1.60 & 2.10 & 0.09 & 0.75 & 1.70 & 1.80 \\
 & World & -- & -- & -- & -- & 0.10 & 0.32 & 2.20 & 2.60 & 0.07 & 0.83 & 1.70 & 1.50 \\
 & Sentiment & 0.13 & 0.24 & 2.50 & 2.90 & 0.12 & 0.27 & 2.50 & 2.80 & 0.12 & 0.64 & 1.90 & 1.90 \\
 & Alphanumeric & 0.17 & -- & 3.00 & -- & 0.15 & -- & 3.00 & -- & 0.17 & -- & 3.00 & -- \\
\cmidrule(lr){2-14}
Qwen3-32B & Ordinal & 0.00 & 0.79 & 1.00 & 1.70 & 0.03 & 0.71 & 1.60 & 2.10 & 0.01 & 0.90 & 1.20 & 1.30 \\
 & World & -- & -- & -- & -- & 0.10 & 0.28 & 2.50 & 2.90 & 0.04 & 0.86 & 1.70 & 1.40 \\
 & Sentiment & 0.09 & 0.29 & 2.20 & 2.80 & 0.14 & 0.29 & 2.40 & 2.70 & 0.10 & 0.34 & 2.20 & 2.40 \\
 & Alphanumeric & 0.16 & -- & 3.00 & -- & 0.15 & -- & 3.00 & -- & 0.16 & -- & 3.00 & -- \\
\bottomrule
\end{tabular}%
}
\caption{Normalised cumulative regret and exploration count at turn 9, by Qwen model and nomenclature. Columns are grouped by domain and split into regret/exploration, with H/M = helpful/misleading framing. Regret is normalised by the same turn-count scale factor used in the notebook. Overall, we found that Qwen3-8B almost never explored, while Qwen3 14B and 32B behaved fairly similarly. To us, this indicated that the 8B model was too small to adequately perform the task. This is particularly clear in its inability to explore in the alphanumeric setting. Since Qwen3 14B and 32B performed similarly, we decided to move forward with the larger model. Scale: \textit{high\_scale}; variance: \textit{No}; history: \textit{Summarized}.}
\label{tab:scaling_comparison_regret_exploration}
\end{table}

\FloatBarrier
\subsection{3 Arm and 5 Arm Comparison}
\label{app:arm_comparison}
\begin{table}[ht]
\centering
\small
\resizebox{\linewidth}{!}{%
\begin{tabular}{llr|rr|rr|rr}
\toprule
Arms & Model & Nom. & Bandit Regret & Bandit Explor & Farm Regret & Farm Explor & Clothing Regret & Clothing Explor \\
\midrule
3 Arms & Qwen3-32B & Helpful & 0.00 & 1.00 & 0.03 & 1.60 & 0.01 & 1.20 \\
 &  & Mislead & 0.79 & 1.70 & 0.71 & 2.10 & 0.90 & 1.30 \\
\midrule
 & Gemini 3.1 Flash Lite & Helpful & 0.13 & 2.80 & 0.15 & 3.00 & 0.15 & 3.00 \\
 &  & Mislead & 0.27 & 2.90 & 0.17 & 3.00 & 0.22 & 2.90 \\
\midrule
 & OLMo-3.1-32B & Helpful & 0.04 & 1.60 & 0.03 & 1.50 & 0.00 & 1.00 \\
 &  & Mislead & 0.79 & 1.50 & 0.91 & 1.10 & 1.00 & 1.00 \\
\midrule
\midrule
5 Arms & Qwen3-32B & Helpful & 0.06 & 1.10 & 0.07 & 1.50 & 0.00 & 1.10 \\
 &  & Mislead & 0.97 & 1.10 & 0.82 & 1.60 & 0.81 & 1.50 \\
\midrule
 & Gemini 3.1 Flash Lite & Helpful & 0.09 & 3.00 & 0.20 & 4.40 & 0.20 & 4.40 \\
 &  & Mislead & 0.60 & 2.67 & 0.29 & 4.86 & 0.34 & 4.33 \\
\midrule
 & OLMo-3.1-32B & Helpful & 0.02 & 1.50 & 0.05 & 2.10 & 0.02 & 1.60 \\
 &  & Mislead & 0.87 & 1.50 & 0.79 & 1.80 & 0.86 & 1.50 \\
\bottomrule
\end{tabular}%
}
\caption{Normalized cumulative regret and exploration count at turn 9 for 3-arm and 5-arm outputs, by model, nomenclature, and domain. H = helpful, M = misleading. Note that the exploration count results are not directly comparable as for the 5 arms, the maximum value is now 5. We find slightly less exploration for Qwen and slightly more exploration for OLMo. We note that Gemini no longer saturates the exploration count metric - this is particularly notable for the abstract bandit, where it does not exceed 3 (out of a maximum of 5). These results indicate that OLMo and Qwen do not necessarily explore more when there are more arms, while Gemini tends to explores more but does not explore fully. We do not observe consistent changes in regret. }
\label{tab:arm_count_comparison}
\end{table}
\FloatBarrier

\subsection{Sample Reasoning Trace}
\label{app:reasoning_trace}
In this section, we provide an example where Gemini follows a misleading semantic prior despite contradictory reward evidence. This example comes from Gemini on the recommendation domain with the misleading world nomenclature, where the user is looking for a shirt and is in a cold climate. The model has previously selected "flannel long-sleeve shirt" for a reward of 25 and "midweight t-shirt" for a reward of 50. It has not selected "tank top" which has a true reward of 75. We observe that the reasoning traces explicitly reject exploring "tank top" due to the negative semantic label. We also observe that the model re-explores "flannel long-sleeve shirt" due to the positive semantic label, even after observing a low reward.

\textbf{Turn 6: The model selects midweight t-shirt as it is the highest observed reward.}

\textit{"To maximize the reward, I need to balance exploring unknown options with exploiting the option that has demonstrated the highest satisfaction score so far. The midweight t-shirt has a solid observed mean of 50.0 after 4 trials. The flannel long-sleeve shirt has a lower observed mean (25.0) and is more appropriate for the "cold" climate, but it has underperformed compared to the t-shirt. The tank top has never been tried and is objectively the worst choice for a "cold" climate, making it a poor candidate for exploration. Given the goal is to maximize reward and the midweight t-shirt is currently yielding the highest return, I will continue to exploit this high-performing option. midweight t-shirt"}

\textbf{Turn 7: The model selects the flannel again despite the lower observed reward signal.}

\textit{"To maximize the reward, I need to balance exploiting the option with the highest observed mean (midweight t-shirt) and exploring the remaining options to ensure no better-performing item is missed, especially given the cold climate context. The "midweight t-shirt" has a high observed mean, but it is not ideally suited for "cold" weather compared to the "flannel long-sleeve shirt." Although the flannel shirt has a lower observed mean based on only two trials, it is theoretically a better fit for the climate, suggesting its true mean satisfaction could be higher. flannel long-sleeve shirt"}

\textbf{Turn 8: The model returns to the midweight t-shirt}

\textit{"To maximize the total reward, I need to balance exploring the items that have not been tested (like the tank top) and exploiting the items that have shown higher satisfaction scores (the midweight t-shirt). Given that the current climate is "cold," a tank top is highly unlikely to provide a high reward. The "midweight t-shirt" has been tested 5 times with a consistent mean of 50.0, which is significantly higher than the "flannel long-sleeve shirt." Continuing to choose the best-performing option in the remaining steps is the optimal strategy. midweight t-shirt "}

\end{document}